\documentclass[10pt,twocolumn,letterpaper]{article}

\usepackage[pagenumbers]{cvpr} 

\usepackage[accsupp]{axessibility}  
\usepackage{pgfplots}
\usepackage{pgfplotstable}
\usepackage{tikz}
\usepackage{graphicx}
\usepackage{multirow}

\usepackage{algorithm}
\usepackage{algorithmic}
\usepackage{listings}

\definecolor{cvprblue}{rgb}{0.21,0.49,0.74}
\usepackage[pagebackref,breaklinks,colorlinks,allcolors=cvprblue]{hyperref}

\def\paperID{8678} 
\def\confName{CVPR}
\def\confYear{2025}

\title{Three Cars Approaching within 100m! Enhancing Distant Geometry by\\ Tri-Axis Voxel Scanning for Camera-based Semantic Scene Completion}

\author{Jongseong Bae$^{1*}$ \hspace{1em}
Junewoo Ha$^{1*}$ \hspace{1em}
Ha Young Kim$^{2\dagger}$\\
$^{1}$Department of Artificial Intelligence, Yonsei University\\
$^{2}$Graduate School of Information, Yonsei University\\
{\tt\small \{js.bae, gkwnsdn0402, hayoung.kim\}@yonsei.ac.kr}
}

\begin{document}
\twocolumn[{
\maketitle
\begin{center}
  \centering
\captionsetup{type=figure}
\vspace{-.5cm}
  \includegraphics[width=1.\textwidth]{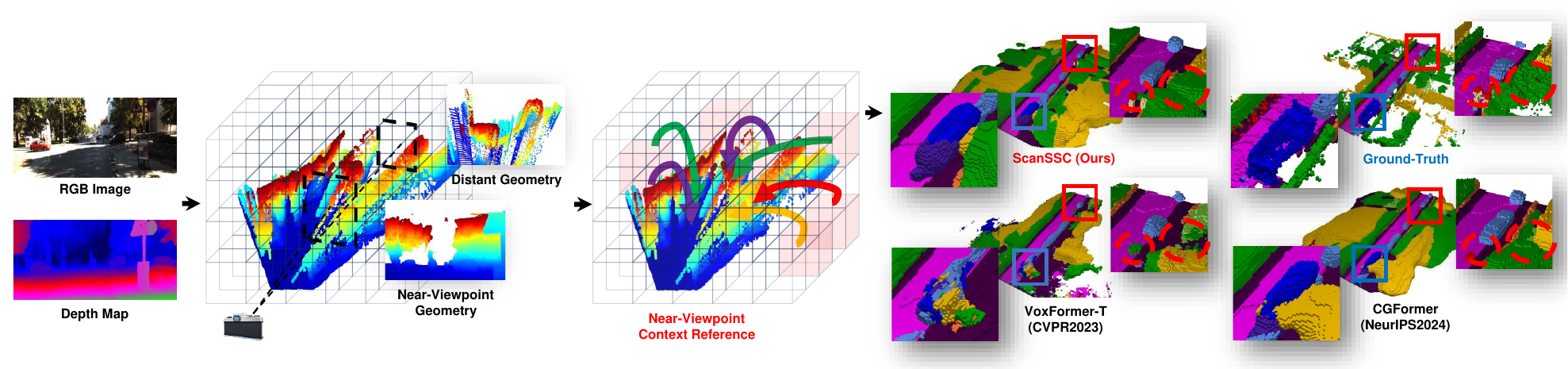}
  
\vspace{-.3cm}
    \captionof{figure}{In camera-based SSC, the projected distant geometry is sparse and unrealistic due to factors such as perspective and occlusion. Our ScanSSC addresses this challenge by referencing distant geometry to the context of more accurate near-viewpoint geometry. As a result, ScanSSC achieves more accurate reconstructions of both distant and near scenes, outperforming existing camera-based SSC methods such as VoxFormer-T~\cite{li2023voxformer} and CGFormer~\cite{yu2024contextgeometryawarevoxel}.}
  \label{fig:concept}
\end{center}}]
\def\thefootnote{*}\footnotetext{Equally contributed}
\def\thefootnote{$\dagger$}\footnotetext{Corresponding author}

\begin{abstract}
\label{sec:abstract}
Camera-based Semantic Scene Completion (SSC) is gaining attentions in the 3D perception field.
However, properties such as perspective and occlusion lead to the underestimation of the geometry in distant regions, posing a critical issue for safety-focused autonomous driving systems.
To tackle this, we propose ScanSSC, a novel camera-based SSC model composed of a Scan Module and Scan Loss, both designed to enhance distant scenes by leveraging context from near-viewpoint scenes.
The Scan Module uses axis-wise masked attention, where each axis employing a near-to-far cascade masking that enables distant voxels to capture relationships with preceding voxels.
In addition, the Scan Loss computes the cross-entropy along each axis between cumulative logits and corresponding class distributions in a near-to-far direction, thereby propagating rich context-aware signals to distant voxels.
Leveraging the synergy between these components, ScanSSC achieves state-of-the-art performance, with IoUs of 44.54 and 48.29, and mIoUs of 17.40 and 20.14 on the SemanticKITTI and SSCBench-KITTI-360 benchmarks.
\end{abstract}
\vspace{-.4cm}
\section{Introduction}
\label{sec:intro}
3D perception of real-world scenes is essential for autonomous driving systems, serving as a cornerstone for navigation and driving safety. Achieving precise reconstruction of the surrounding geometry is critical but challenging due to the geometric discrepancies between sensor data and real-world coordinates.

3D semantic scene completion (SSC) is a recently proposed~\cite{song2017semantic} task that jointly predicts the 3D geometry and semantics of the surrounding scene.
Since the release of the SemanticKITTI benchmark~\cite{behley2019semantickitti}, numerous studies have explored outdoor SSC.
While LiDAR-based methods~\cite{xia2023scpnet,cheng2021s3cnet,yan2021sparse,yang2021semantic} remain the primary approach due to their strong performance, they come with the high cost of LiDAR sensors.
Recently, since MonoScene~\cite{cao2022monoscene} initially tackled monocular SSC, camera-based methods~\cite{li2023voxformer,huang2023tri,zhang2023occformer,miao2023occdepth,jiang2024symphonize,yu2024contextgeometryawarevoxel,li2024hierarchical} have gained the spotlight owing to their rich visual information and cost-effectiveness.

In recent camera-based SSC approaches, techniques such as Features Line-of-Sight Projection (FLoSP)~\cite{cao2022monoscene}, back projection using 2D depth estimation~\cite{li2023voxformer,jiang2024symphonize}, and LSS~\cite{philion2020lift} feature volume~\cite{yu2024contextgeometryawarevoxel,li2024hierarchical} have been employed for lifting 2D features.
However, these methods inherit common limitations of camera images, such as perspective and occlusion, resulting in sparse and unreliable geometry projections for distant views compared to closer ones.
Our analysis in Sec.~\ref{sec:method_pre} demonstrates that this issue adversely impacts SSC performance, as mIoU values significantly decrease with increasing distance from the viewpoint. At the same time, IoU and recall metrics exhibit similar declining trends.
This circumstance can be interpreted as an underestimation of geometry, which could pose critical real-world challenges for driving safety.

In this paper, we aim to enhance distant geometry by guiding it with the context of finely constructed near-viewpoint geometry (Fig.~\ref{fig:concept}).
We propose ScanSSC, a novel camera-based SSC model composed of two main components: Scan Module and Scan Loss.
Scan Module employs three axis-wise masked self-attentions within an autoregressive Transformer framework~\cite{vaswani2023attentionneed}.
Each axis applies axis-specific masking, enabling distant voxels to reference a broad range of prior voxels while preventing close voxels from being influenced by uncertain, occluded voxels behind them. 
Additionally, we propose Scan Loss, which computes the cross-entropy of cumulatively averaged logits, spreading from distant voxels to close ones, along with the corresponding accumulated class distributions for each axis.
By repeatedly including the logits of distant voxels in various regional loss calculations, Scan Loss effectively propagates rich contextual information to these distant voxels.

Through extensive experiments, we observe an impressive synergy between the Scan Module and Scan Loss.
Leveraging this synergy, ScanSSC achieves significantly improved SSC results, demonstrating robust performance across varying distances from viewpoints.
As a result, ScanSSC markedly outperforms previous methods, achieving state-of-the-art (SOTA) IoU and mIoU scores of 44.54 and 48.29, and 17.40 and 20.14 on the SemanticKITTI and SSCBench-KITTI-360 benchmarks~\cite{li2024sscbenchlargescale3dsemantic}.

Our contributions are summarized as follows:
\begin{itemize}
    \item We first unveil the issue of distance-dependent completion imbalance in camera-based SSC through a comprehensive analysis of existing methods.
    \item Based on the analysis, specifically, to enhance the distant geometry, we design the Scan Module, which employs axis-wise masked self-attention, enabling distant voxels to be refined by the context of preceding voxels.
    \item We also propose the Scan Loss, defined as cross-entropy between the cumulative average of prediction logits and accumulated class distributions, designed to propagate abundant contextual signals to distant voxels.
    \item By incorporating Scan Module and Scan Loss, we introduce a novel camera-based SSC model, ScanSSC. Leveraging the synergy of both components, ScanSSC achieves SOTA IoUs of 44.54 and 48.29, and mIoUs of 17.40 and 20.14 on the SemanticKITTI and SSCBench-KITTI-360 benchmarks.
\end{itemize}
\section{Related Work}
\label{sec:related_work}
\textbf{Camera-Based 3D Perception.} 
The primary tasks in 3D perception include 3D object detection (OD) and bird's-eye-view (BEV) segmentation.
Inspired by DETR~\cite{carion2020end} and its family~\cite{wang2022detr3d, meng2021conditional, wang2022anchor, liu2022dab} of 2D OD models, methods such as DETR3D~\cite{wang2022detr3d}, PETR~\cite{liu2022petr}, and PETRv2~\cite{liu2023petrv2} have been proposed, utilizing object queries and camera projection matrices. DETR3D~\cite{wang2022detr3d} and PETR~\cite{liu2022petr} connect 2D features with 3D space using object queries, eliminating the need for post-processing techniques like NMS~\cite{hosang2017learning} when predicting bounding boxes and labels.
PETRv2~\cite{liu2023petrv2} enhances the original version by incorporating temporal information through temporal alignment.
In BEV segmentation, pixel-wise depth distribution is used to convert camera images into 3D point cloud features projected onto the BEV plane, as demonstrated by methods such as LSS~\cite{philion2020lift} and FIERY~\cite{hu2021fiery}.
Additionally, BEVFormer~\cite{li2022bevformer} utilizes pre-defined BEV queries and integrates spatiotemporal features through an attention mechanism.

The camera-based approach is cost-effective and easy to implement, making it more suitable for autonomous driving systems requiring real-time situational awareness than the cost-heavy LiDAR-based perception.
Therefore, we propose a model that understands the holistic scene through camera-based perception.\\

\noindent\textbf{3D Semantic Scene Completion.} 
3D SSC involves voxelizing a scene by predicting the occupancy and semantics of each voxel.
Since SSCNet~\cite{song2017semantic} introduced SSC, various approaches using LiDAR point clouds and camera images have emerged.
Point-based methods~\cite{xia2023scpnet,cheng2021s3cnet,yan2021sparse,yang2021semantic} achieve high performance due to accurate depth data but are computationally expensive.
Camera-based SSC~\cite{cao2022monoscene, li2023voxformer, zheng2024monoocc, jiang2024symphonize, yu2024contextgeometryawarevoxel} requires lifting 2D features into 3D.
MonoScene~\cite{cao2022monoscene} connects 2D and 3D UNets via FLoSP, sparking further camera-based SSC research.
VoxFormer~\cite{li2023voxformer} employs a two-stage approach with depth-based occupancy prediction followed by semantic prediction.
Subsequent research has explored geometric information from depth maps, notably MonoOCC~\cite{zheng2024monoocc} with a large pre-trained backbone and Symphonies~\cite{jiang2024symphonize}, which uses instance queries for enhanced instance prediction. However, projecting 3D to 2D space can lead to overlapping 2D points from different 3D locations. To address this, CGFormer~\cite{yu2024contextgeometryawarevoxel} applies LSS~\cite{philion2020lift} to generate point cloud features from 2D features and depth probability, using dependent voxel queries to capture unique image characteristics. Nevertheless, monocular methods face the inherent limitation of decreasing depth accuracy with distance.

To address these issues, we propose a model incorporating the Scan Module applying masked self-attention with axis-specific masks and the Scan Loss function, which extends cross-entropy loss~\cite{zhang2018generalized}.

\section{Method}
\label{sec:method}
This section analyzes the issues in existing camera-based 3D SSC methods and describes our proposed model, ScanSSC.
In Sec.~\ref{sec:method_pre}, we analyze the prediction results of VoxFormer~\cite{li2023voxformer}, a milestone in camera-based 3D SSC methods, to identify key issues.
Sec.~\ref{sec:method_overview} presents an overview of the proposed ScanSSC architecture, while Sec.~\ref{sec:method_module} and Sec.~\ref{sec:method_loss} detail the Scan Module and Scan Loss, each designed to address the identified issues.
Finally, Sec.~\ref{sec:method_tr_loss} covers the overall training loss.

\subsection{Preliminary: Distance-Dependent Completion Imbalance in Camera-Based SSC}
\label{sec:method_pre}
\vspace{-.4cm}
\begin{figure}[ht]
    \centering
    \begin{subfigure}{0.32\linewidth}
        \subfloat[Depth-axis]{\includegraphics[width=\linewidth]{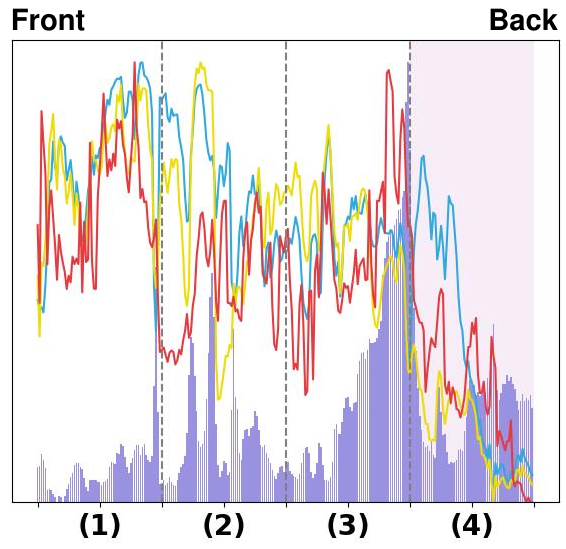}}
        \label{fig:depth_wise}
    \end{subfigure}
    \hspace{0.00\linewidth}
    \begin{subfigure}{0.32\linewidth}
        \subfloat[Width-axis]{\includegraphics[width=\linewidth]{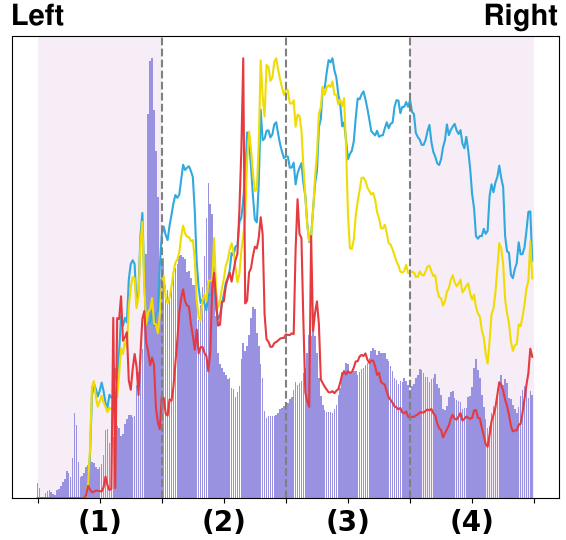}}
        \label{fig:width_wise}
    \end{subfigure}
    \hspace{0.00\linewidth}
    \begin{subfigure}{0.32\linewidth}
        \subfloat[Height-axis]{\includegraphics[width=\linewidth]{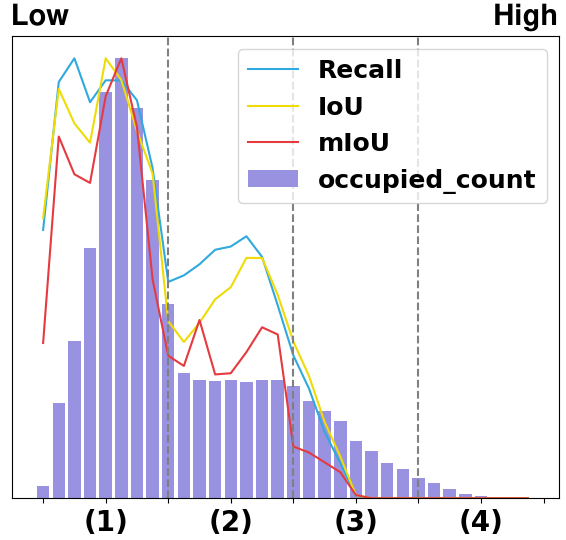}}
        \label{fig:height_wise}
    \end{subfigure}
    
\vspace{-.2cm}
    \caption{Axis-wise trends in recall, IoU, and mIoU for VoxFormer~\cite{li2023voxformer}, along with the ground-truth occupied voxel distributions, on the SemanticKITTI~\cite{behley2019semantickitti} validation data. All figures are presented in a graph, scaled between 0 and 1. Each graph is binned with sizes of 256, 256, and 32 for the depth, width, and height axes, respectively. (1) to (4) represent each segment when the axis is divided into four equal parts.
    }
    \label{fig:preliminary}
\end{figure}
\vspace{-0.6cm}
\begin{table}[h]
\centering
\resizebox{\columnwidth}{!}{%
\begin{tabular}{c|ccc|ccc|ccc}
\hline
 &
  \multicolumn{3}{c|}{Depth} &
  \multicolumn{3}{c|}{Width} &
  \multicolumn{3}{c}{Height} \\ \cline{2-10} 
 &
  Recall &
  IoU &
  mIoU &
  Recall &
  IoU &
  mIoU &
  Recall &
  IoU &
  mIoU \\ \hline
(1) &
  78.18 &
  60.55 &
  18.21 &
  \cellcolor[HTML]{F6EDF6}\textbf{20.04} &
  \cellcolor[HTML]{F6EDF6}\textbf{15.37} &
  \cellcolor[HTML]{F6EDF6}\textbf{1.71} &
  79.04 &
  54.43 &
  11.40 \\
(2) &
  73.35 &
  53.66 &
  14.32 &
  60.83 &
  49.10 &
  5.75 &
  4.79 &
  29.38 &
  5.21 \\
(3) &
  63.46 &
  51.52 &
  16.10 &
  78.56 &
  56.49 &
  4.16 &
  9.13 &
  7.34 &
  0.72 \\
(4) &
  \cellcolor[HTML]{F6EDF6}{\textbf{42.42}} &
  \cellcolor[HTML]{F6EDF6}{\textbf{21.47}} &
  \cellcolor[HTML]{F6EDF6}{\textbf{8.98}} &
  \cellcolor[HTML]{F6EDF6}{\textbf{67.60}} &
  \cellcolor[HTML]{F6EDF6}{\textbf{35.73}} &
  \cellcolor[HTML]{F6EDF6}{\textbf{2.58}} &
  0.00 &
  0.00 &
  0.00 \\ \hline
\end{tabular}%
}
\vspace{-0.2cm}
\caption{Recall, IoU, and mIoU values for VoxFormer~\cite{li2023voxformer}, calculated by dividing each axis into 4 intervals, as labeled in (1) to (4).}
\vspace{-0.2cm}
\label{tab:analysis_axiswise}
\end{table}

\noindent Camera-based SSC methods~\cite{cao2022monoscene, li2023voxformer, zheng2024monoocc, jiang2024symphonize, yu2024contextgeometryawarevoxel} utilize the rich visual information from RGB images but are vulnerable to inaccurate depth information and the effects of occlusion.
Although previous methods~\cite{li2023voxformer,yu2024contextgeometryawarevoxel} have acknowledged this issue, they have yet to focus on addressing it comprehensively.
In this subsection, we systematically analyze this challenge based on the prediction results of VoxFormer~\cite{li2023voxformer}. 
Fig.~\ref{fig:preliminary} graphically presents the axis-wise trends of recall, IoU, and mIoU metrics of VoxFormer, along with the distributions of occupied ground-truth (GT) voxels.
Additionally, Tab.~\ref{tab:analysis_axiswise} provides the numerical values averaged over four sequentially divided intervals (1)-(4).\\

\noindent\textbf{Depth-Axis Analysis.}
In Fig.~\ref{fig:preliminary}\subref{fig:depth_wise}, we observe a trend where, despite a comparable number of occupied GT voxels in the backside section (4) to those in the frontside sections (1) and (2), the recall, mIoU, and IoU values decrease as depth increases.
Tab.~\ref{tab:analysis_axiswise} provides numerical evidence indicating that the average scores decrease with distance: the three metrics for the far area (4) are approximately 35.76, 39.08, and 9.23 lower than those for the near area (1), respectively. 
We deduce that one of the primary factors contributing to this is the sparsity of projected geometry resulting from perspective and occlusion. 
Furthermore, inaccuracies in depth estimation for distant points may also influence this situation.\\

\noindent\textbf{Width-Axis Analysis.}
For width, Fig.~\ref{fig:preliminary}\subref{fig:width_wise} shows a trend where three metrics decrease as we move from the center toward the outer areas (1) and (4), different from the distribution of occupied GT voxels.
According to the values in Tab.~\ref{tab:analysis_axiswise}, the area (1) is approximately 49.66, 37.43, and 3.25 lower than the average of the middle areas (2) and (3), while the area (4) is around 2.1, 17.07, and 2.38 lower. 
This phenomenon can be explained by the perspective of the camera's view frustum, which follows a conical shape, causing the side areas to have less visual information compared to the center points. 
Additionally, in a driving context, it is important to note that the side areas are more vulnerable to occlusion than the direct line of sight.\\

\noindent\textbf{Height-Axis Analysis.}
For height, Fig.~\ref{fig:preliminary}\subref{fig:height_wise} shows a trend where values decrease as we move from the lower area (1) to the higher area (4), aligning with the distribution trend of occupied voxels.
Notably, Tab.~\ref{tab:analysis_axiswise} shows that all values are zero in the higher area (4). 
This is challenging to interpret as a side effect of camera-based 3D reconstruction; rather, it is likely due to voxel distribution, with most objects concentrated at lower levels and higher regions primarily empty.\\

From these analyses, we gain the insight that the prediction accuracy of current camera-based SSC methods tends to decrease as the distance from the viewpoint increases.
To tackle these challenges, we introduce ScanSSC, designed to execute near-to-far geometric refinement in an axis-wise manner. 
For clarity, we define the terms as follows: ``near-to-far" refers to front-to-back along the depth axis, center-to-side along the width axis, and high-to-low along the height axis.
For the height axis, a BEV serves as the reference for the definition, as the complexity of GT geometry is inversely related to height; lower positions are more challenging to complete due to a greater concentration of object voxels compared to higher positions.

\subsection{Overview}
\label{sec:method_overview}
\begin{figure*}[!t]
    \centering
    \includegraphics[width=1.\linewidth]{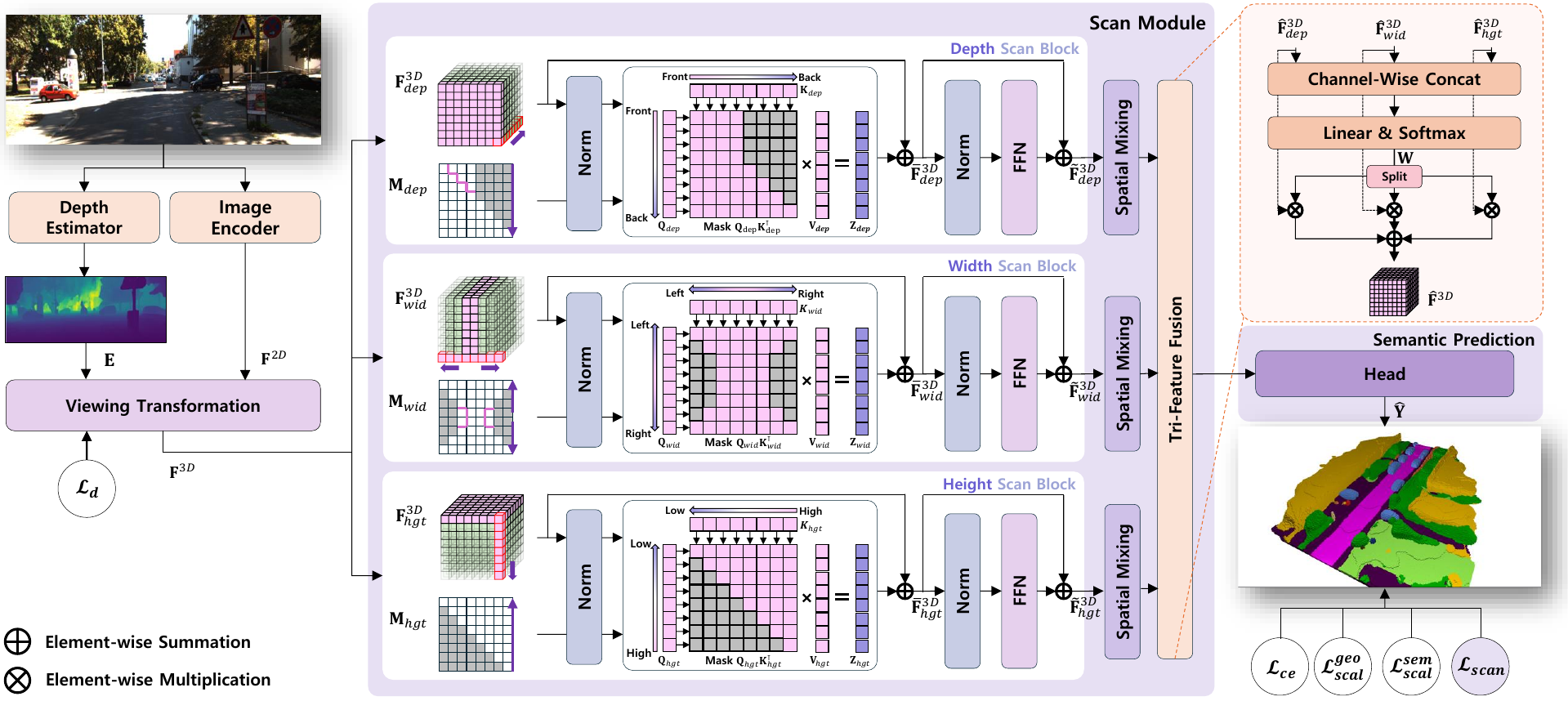}
    \vspace{-0.6cm}
    \caption{The overall architecture of the proposed ScanSSC. After ${\textbf{F}}^{3D}$ is obtained through the viewing transformation, it is passed through the three parallel Scan blocks of the Scan Module. Each block performs masked self-attention along the axis highlighted in red. The purple arrows indicate the 'near-to-far' direction, implemented by the corresponding mask below. ${\textbf{Q}}_{axis}$, ${\textbf{K}}_{axis}$, $\textbf{V}_{axis}$, and $\textbf{Z}_{axis}$ denote the query, key, value, and output features of attention, respectively, where ${axis} \in \{{dep}, {wid}, {hgt}\}$.}
    \label{fig:enter-label}
\vspace{-0.3cm}

\end{figure*}
The overall architecture of ScanSSC is depicted in Fig.~\ref{fig:enter-label}.
ScanSSC comprises three subparts: viewing transformation, Scan Module, and semantic prediction.\\

\noindent\textbf{Viewing Transformation.} We follow the CGFormer~\cite{yu2024contextgeometryawarevoxel} method for viewing transformation, which combines the LSS~\cite{philion2020lift} feature volume with a depth-based query proposal method~\cite{li2023voxformer,jiang2024symphonize}.
The process is briefly outlined as follows.

Given a monocular RGB image $\textbf{I}\in\mathbb{R}^{H\times W\times 3}$, where $(H, W)$ denotes the image resolution, the image feature $\text{\textbf{F}}^{2D}\in\mathbb{R}^{H'\times W'\times C}$ and the depth map $\text{\textbf{E}}\in\mathbb{R}^{H\times W}$ are extracted through the image encoder and depth estimator, respectively.
Here, ($H'$, $W'$) and $C$ denote the resolution and channel of the feature, respectively.
Using $\text{\textbf{F}}^{2D}$ and $\text{\textbf{E}}$, we obtain the LSS volume $\text{\textbf{F}}\in\mathbb{R}^{H'\times W'\times D\times C}$ by taking the outer product of $\text{\textbf{F}}^{2D}$ and the depth probability $\text{\textbf{D}}\in\mathbb{R}^{H'\times W'\times D}$, which is extracted from an additional depth network.
Here, $D$ refers to the discretized depth bins.
Next, \textbf{F} is projected onto the voxel proposal of initial voxel grid $\textbf{P}\in\mathbb{R}^{\hat{X}\times \hat{Y}\times \hat{Z}\times C}$ using deformable cross-attention~\cite{zhu2021deformabledetrdeformabletransformers}. After passing through deformable self-attention, the 3D-lifted feature $\text{\textbf{F}}^{3D}\in \mathbb{R}^{\hat{X}\times \hat{Y}\times \hat{Z}\times C}$ is obtained.
$\hat{X}$, $\hat{Y}$, and $\hat{Z}$ denote the depth, width, and height of the voxel grid, respectively.\\

\noindent\textbf{Scan Module.} As mentioned in Sec.~\ref{sec:intro}, we aim to refine distant geometry by guiding it with the more accurate context of near-viewpoint geometry.
Thus, ScanSSC sequentially scans the voxelized feature along each axis in a near-to-far direction through the Scan Module.
Given $\text{\textbf{F}}^{{3D}}$, the scanned voxel feature $\hat{\text{\textbf{F}}}^{{3D}}\in \mathbb{R}^{\hat{X}\times \hat{Y}\times \hat{Z}\times C}$ is derived as follows:
{\small
\begin{equation}
    \hat{\text{\textbf{F}}}^{{3D}}=\text{ScanModule}(\text{\textbf{F}}^{{3D}}).
\end{equation}}
Details of the Scan Module are explained in Sec.~\ref{sec:method_module}.

\noindent\textbf{Semantic Prediction.} $\hat{\text{\textbf{F}}}^{{3D}}$ is then passed to the prediction head, which consists of a lightweight 3D convolutional network with a 3$\times$3 kernel, followed by normalization and a linear projection.
The output feature is subsequently upscaled via trilinear interpolation to align with the target voxel grid.
In short, the prediction logit $\hat{\text{\textbf{Y}}}\in \mathbb{R}^{{X}\times {Y}\times {Z}\times P}$ is computed as:
{\small
\begin{equation}
    \hat{\text{\textbf{Y}}}=\text{Upsample}(\text{Linear}(\text{Norm}(\text{Conv3D}(\hat{\text{\textbf{F}}}^{{3D}})))),
\end{equation}}
where (${X}, {Y}, {Z}$) denotes the spatial dimensions of the target voxel grid, and $P$ is the number of semantic classes.

\subsection{Scan Module}
\label{sec:method_module}
The Scan Module is illustrated in Fig.~\ref{fig:enter-label}.
Building on the analysis in Sec.~\ref{sec:method_pre}, our goal is to enhance distant voxel features by leveraging the more established features of near-viewpoint voxels along each axis.
To achieve this, we develop the Scan Module, which is divided into three branches, each focusing on a specific axis and employing an appropriate strategy.
The module includes three parallel axis-specific Scan blocks, each followed by a spatial mixing network and a fusion process for the three parallel features.
We provide a detailed explanation of each component.\\

\noindent\textbf{Scan Block.}
Like modern vision Transformer blocks~\cite{liu2021swintransformerhierarchicalvision,yu2022metaformeractuallyneedvision,Yu_2024}, the Scan block is composed of a sequence of self-attention (SA) and feed-forward network (FFN) subblocks, each featuring pre-normalization and a residual connection. 
For the SA layer, the Scan block utilizes masked self-attention~\cite{vaswani2023attentionneed} with an axis-specific mask $\textbf{M}_{axis}\in \{\textbf{M}_{dep}, {\textbf{M}}_{wid}, {\textbf{M}}_{hgt}\}$, where each mask corresponds to the depth, width, and height axes, respectively. 
We specifically organize the masking positions of $\textbf{M}_{dep}, \textbf{M}_{wid}$, and $\textbf{M}_{hgt}$ based on the analysis in Sec.~\ref{sec:method_pre}.

$\textbf{M}_{dep}, \textbf{M}_{wid}$, and $\textbf{M}_{hgt}$ are designed to enable distant voxel features to reference preceding voxel features while simultaneously preventing the reverse, thereby minimizing the influence of inaccurate features from distant voxels on previous voxels for each axis.
Consequently, we apply a cascading mask that enables attention computation for preceding voxels while blocking subsequent voxels.
Since the indices for the depth, width, and height axes of the voxel grid are arranged from front to back, left to right, and bottom to top, $\textbf{M}_{dep}$ is configured as an upper triangular matrix with zeros along the diagonal.
In contrast, $\textbf{M}_{wid}$ is designed in an hourglass shape, while $\textbf{M}_{hgt}$ resembles a lower triangular matrix with a zero diagonal.
Additionally, we define a margin region within a specific range of well-established near-viewpoint voxels, where masking is removed to allow unrestricted interactions.
For $\textbf{M}_{dep}$, $\textbf{M}_{wid}$, and $\textbf{M}_{hgt}$, the margin regions are by default set from the start of ``near-to-far'' of each axis as follows: 50\% backward along the depth axis, 25\% to each side (for a total of 50\%) along the width axis, and 0\% along the height axis.

The 3D feature $\textbf{F}^{3D}$, obtained through viewing transformation, is separated into three different flattened features: $\textbf{F}^{3D}_{dep}\in \mathbb{R}^{(\hat{Y}\hat{Z})\times \hat{X}\times C}$, $\textbf{F}^{3D}_{wid}\in \mathbb{R}^{(\hat{X}\hat{Z})\times \hat{Y}\times C}$, and $\textbf{F}^{3D}_{hgt}\in \mathbb{R}^{(\hat{X}\hat{Y})\times \hat{Z}\times C}$.
Each of the three features is individually input into a corresponding Scan block along with the respective attention masks $\textbf{M}_{dep}, \textbf{M}_{wid}$, and $\textbf{M}_{hgt}$.
Each per-axis Scan block operates as follows:
{\small
\begin{align}
\begin{split}
    &\mathbf{\bar{F}}^{3D}_{axis} = \mathbf{F}^{3D}_{axis} + \text{MaskedSA}(\text{Norm}_1(\mathbf{F}^{3D}_{axis}), \mathbf{M}_{axis}),\\
    &\mathbf{\tilde{F}}^{3D}_{axis} = \mathbf{\bar{F}}^{3D}_{axis} + \text{FFN}(\text{Norm}_2(\mathbf{\bar{F}}^{3D}_{axis})),
\end{split}
\end{align}}
where ${axis}$ is an element of the set $\{{dep}, {wid}, {hgt}\}$.
For both $\text{Norm}_1(\cdot)$ and $\text{Norm}_2(\cdot)$, we use layer normalization~\cite{ba2016layernormalization}. $\text{MaskedSA}(\cdot)$ refers to the masked self-attention layer, and $\text{FFN}(\cdot)$ denotes a 2-layer FFN with a ReLU~\cite{agarap2019deeplearningusingrectified} activation function.
\\

\noindent\textbf{Spatial Mixing Network.} 
This network is designed to enhance regional spatial patterns in each scan feature, $\mathbf{\tilde{F}}^{3D}_{dep}$, $\mathbf{\tilde{F}}^{3D}_{wid}$, and $\mathbf{\tilde{F}}^{3D}_{hgt}$.
It first uses a lightweight ResNet~\cite{he2015deepresiduallearningimage} to extract multi-scale features, which are subsequently fused using a 3D Feature Pyramid Network (FPN)~\cite{lin2017feature}.
The output features $\mathbf{\hat{F}}^{3D}_{dep}$, $\mathbf{\hat{F}}^{3D}_{wid}$, and $\mathbf{\hat{F}}^{3D}_{hgt}$ are respectively obtained through the per-axis spatial mixing network, which operates as follows:
{\small
\begin{equation}
    \mathbf{\hat{F}}^{3D}_{axis} = \text{SMN}(\mathbf{\tilde{F}}^{3D}_{axis}),
\end{equation}}
where ${axis}\in \{{dep}, {wid}, {hgt}\}$, and $\text{SMN}(\cdot)$ represents the spatial mixing network.\\

\noindent\textbf{Tri-Feature Fusion.}
Finally, the three axis-wise features $\mathbf{\hat{F}}^{3D}_{{dep}}$, $\mathbf{\hat{F}}^{3D}_{{wid}}$, and $\mathbf{\hat{F}}^{3D}_{{hgt}}$ are fused by a weighted summation.
\begin{figure}[!t]
    \centering
    \includegraphics[width=1\linewidth]{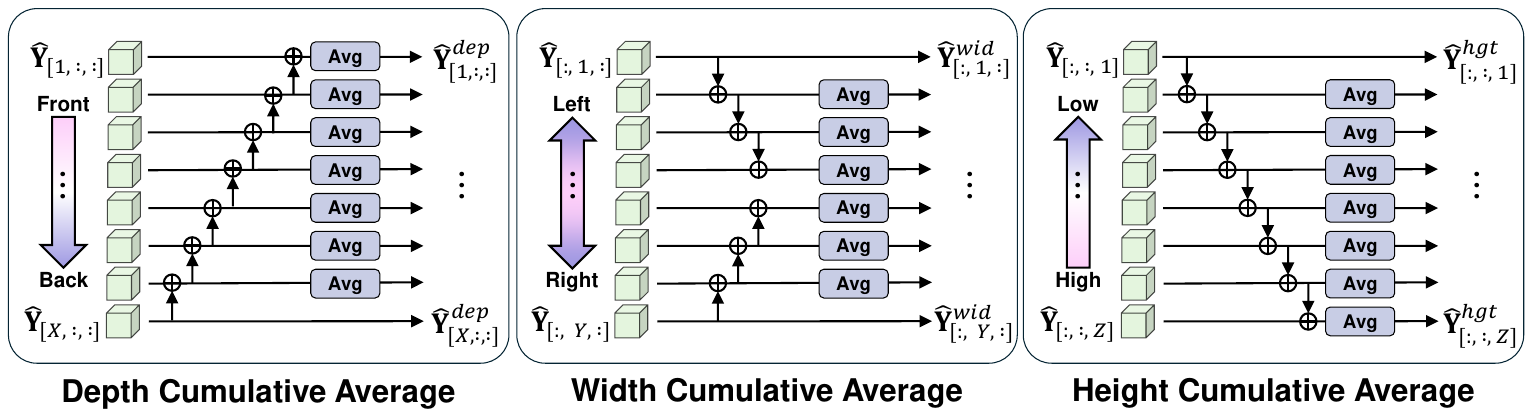}
    \vspace{-0.6cm}
    \caption{Visual overview of the axis-wise cumulative voxel averaging in Scan Loss. Each voxel represents a predicted class logit.}
    \label{fig:loss_fig}
\vspace{-0.4cm}
\end{figure}

To calculate the voxel-wise weight values, $\textbf{W}\in\mathbb{R}^{\hat{X}\times \hat{Y}\times \hat{Z}\times 3}$, these features are concatenated along the channel dimension, followed by a linear projection and then a softmax function, as shown below:
{\small
\begin{equation}
    \textbf{W}=\text{Softmax}(\text{Linear}(\text{Concat}(\mathbf{\hat{F}}^{3D}_{dep}, \mathbf{\hat{F}}^{3D}_{wid}, \mathbf{\hat{F}}^{3D}_{hgt}))).
\end{equation}}
Using these weights, the integrated voxel feature $\mathbf{\hat{F}}^{3D}\in\mathbb{R}^{\hat{X}\times \hat{Y}\times \hat{Z}\times C}$ is calculated as follows: 
{\small
\begin{equation}
    \mathbf{\hat{F}}^{3D}=\sum_{axis}^{\{dep, wid, hgt\}}\mathbf{W}_{[:,:,:,axis]}\otimes\mathbf{\hat{F}}^{3D}_{axis},
\end{equation}}
where $\otimes$ represents element-wise multiplication.

\subsection{Scan Loss}
\label{sec:method_loss}
In addition to the Scan Module, we propose the Scan Loss, $\mathcal{L}_{scan}$, to achieve near-to-far geometric refinement at the training level. 
Previous methods~\cite{li2023voxformer,jiang2024symphonize,yu2024contextgeometryawarevoxel} have primarily used voxel-wise cross-entropy loss~\cite{zhang2018generalized}, which does not account for relationships between multiple voxels or the geometric distribution of neighboring voxels.

$\mathcal{L}_{scan}$ is designed to enhance the training of distant voxels by incorporating their relational information with well-established near-viewpoint voxels.
It accomplishes this by calculating the cross-entropy using their averaged class logits.
Similar to the Scan Module, which cascadingly reflects the relationship between a distant voxel and preceding voxels, Scan Loss uses cumulatively averaged logits along a particular axis, as illustrated in Fig.~\ref{fig:loss_fig}.
In this approach, cumulative averaging begins from distant voxels, allowing them to learn more effectively through exposure to diverse loss calculations and various geometric distributions.
For closer voxels, only the overall class distribution is used.
As a result, Scan Loss enables distant voxels to gain richer, context-aware representations from earlier loss calculations.
Based on the voxel grid axes, $\mathcal{L}_{scan}$ is categorized into $\mathcal{L}_{scan}^{{dep}}$, $\mathcal{L}_{scan}^{{wid}}$, and $\mathcal{L}_{scan}^{{hgt}}$.
Unlike the Scan Module, the cumulative averaging is applied distant voxels toward closer voxels along each axis, thereby propagating richer contextual signals to distant voxels—specifically from back to front for the depth axis, side to center for the width axis, and bottom to top for the height axis.

Given the logit feature $\hat{\text{\textbf{Y}}}$, axis-wise cumulatively averaged logit features ($\hat{\text{\textbf{Y}}}^{dep}$, $\hat{\text{\textbf{Y}}}^{wid}$, $\hat{\text{\textbf{Y}}}^{hgt}$) are calculated as:
\vspace{-.35cm}
{\small
\begin{align}
\begin{split}
    &\hat{\text{\textbf{Y}}}_{[x, :, :]}^{dep}=\frac{1}{X-x+1}\sum^{X}_{i=x}{\hat{\text{\textbf{Y}}}_{[i, :, :]}},\\
    &\hat{\text{\textbf{Y}}}_{[:, y, :]}^{wid}=\begin{cases}
    \frac{1}{y}\sum^{y}_{j=1}{\hat{\text{\textbf{Y}}}_{[:, j, :]}}, & y\le \frac{Y}{2}\\
    \frac{1}{Y-y+1}\sum^{Y}_{j=y}{\hat{\text{\textbf{Y}}}_{[:, j, :]}}, & y> \frac{Y}{2}\end{cases},\\
    &\hat{\text{\textbf{Y}}}_{[:, :, z]}^{hgt}=\frac{1}{z}\sum^{z}_{k=1}{\hat{\text{\textbf{Y}}}_{[:, :, k]}},
\end{split}
\label{eq:7}
\end{align}}
where $\hat{\text{\textbf{Y}}}_{[i,j,k]}$ denotes the $i$-th, $j$-th, and $k$-th element of $\hat{\text{\textbf{Y}}}$ for depth, width, and height axes.
The corresponding target voxels ($\textbf{\text{Y}}^{dep}$, $\textbf{\text{Y}}^{wid}$, $\textbf{\text{Y}}^{hgt}$) are calculated from the GT voxel grid \textbf{Y} in the same way as in Eq.~\ref{eq:7}.
With those logits and targets, $\mathcal{L}_{scan}^{dep}$, $\mathcal{L}_{scan}^{wid}$, and $\mathcal{L}_{scan}^{hgt}$ are computed as follows:
{\small
\begin{equation}
    \mathcal{L}_{scan}^{axis}=\text{CE}(\hat{\text{\textbf{Y}}}^{axis},\textbf{Y}^{axis})\,\,\text{for}\,\,{axis} \in \{dep,wid,hgt\},
\end{equation}}
where CE($\cdot$,$\cdot$) refers to the cross-entropy function.
Finally, $\mathcal{L}_{scan}$ is represented as the summation of three losses as:
{
\small
\begin{equation}
    \mathcal{L}_{scan} = \mathcal{L}_{scan}^{dep} + \mathcal{L}_{scan}^{wid} + \mathcal{L}_{scan}^{hgt}.
\end{equation}}

\subsection{Training Strategy}
\label{sec:method_tr_loss}
Previous works~\cite{cao2022monoscene,li2023voxformer,jiang2024symphonize} have commonly utilized cross-entropy loss $\mathcal{L}_{ce}$, affinity losses $\mathcal{L}_{scal}^{geo}$ and $\mathcal{L}_{scal}^{sem}$~\cite{cao2022monoscene}, and especially depth loss $\mathcal{L}_{d}$~\cite{yu2024contextgeometryawarevoxel} for viewing transformation in CGFormer.
We also employ these losses along with our $\mathcal{L}_{scan}$, hence, the total loss $\mathcal{L}$ is as follows:
{
\small
\begin{equation}
    \mathcal{L}=\mathcal{L}_{ce}+\mathcal{L}_{scal}^{geo}+\mathcal{L}_{scal}^{sem}+\lambda_{d}\mathcal{L}_{d}+\lambda_{scan}\mathcal{L}_{scan},
\end{equation}}
where we set $\lambda_d$ and $\lambda_{scan}$ to 0.001 and 1, respectively.

\section{Experiments}
\label{sec:experiment}
\subsection{Quantitative Results}
\vspace{-0.6cm}
\begin{figure}[hbt!]
    \centering
    \begin{subfigure}{0.32\linewidth}
        \subfloat[Depth-axis]{\includegraphics[width=\linewidth]{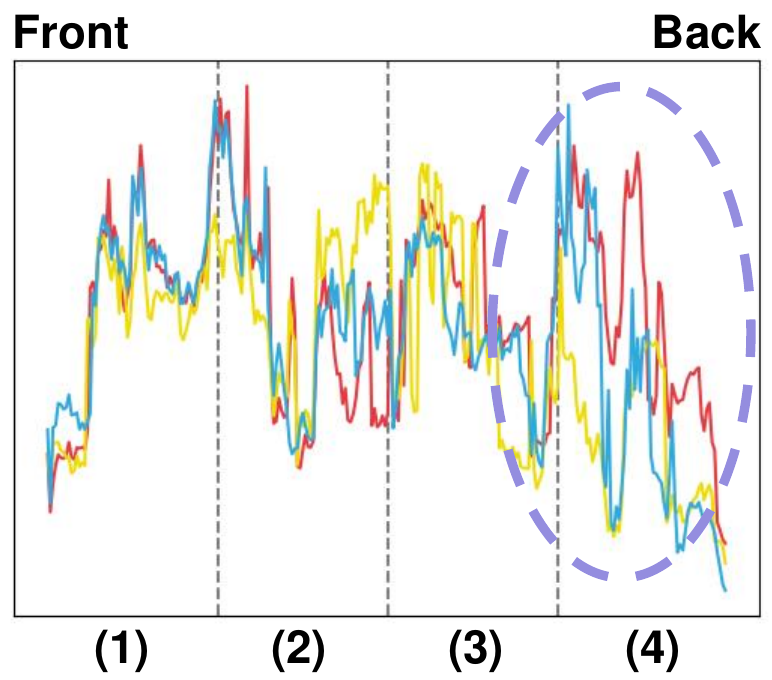}}
        \label{fig:depth_compared}
    \end{subfigure}
    \hspace{0.00\linewidth}
    \begin{subfigure}{0.32\linewidth}
        \subfloat[Width-axis]{\includegraphics[width=\linewidth]{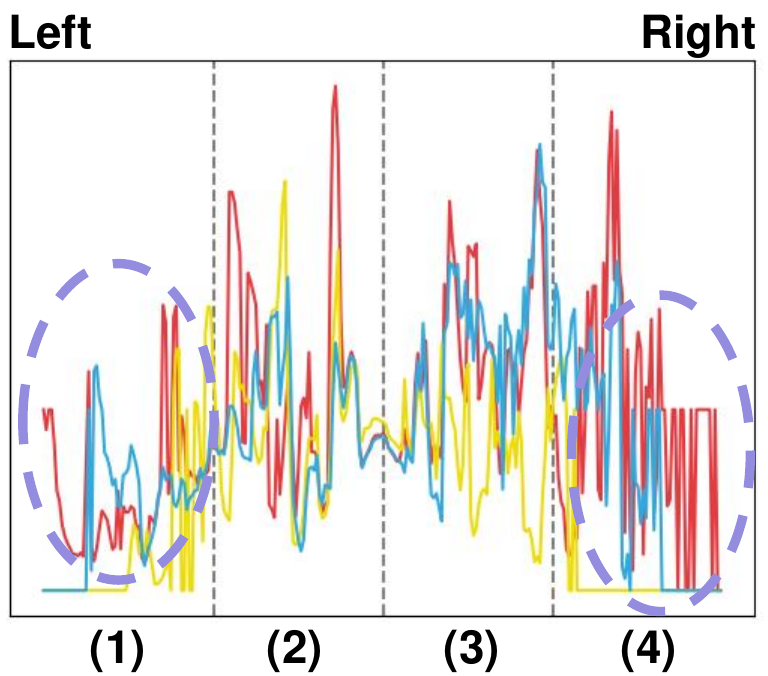}}
        \label{fig:width_compared}
    \end{subfigure}
    \hspace{0.00\linewidth}
    \begin{subfigure}{0.32\linewidth}
        \subfloat[Height-axis]{\includegraphics[width=\linewidth]{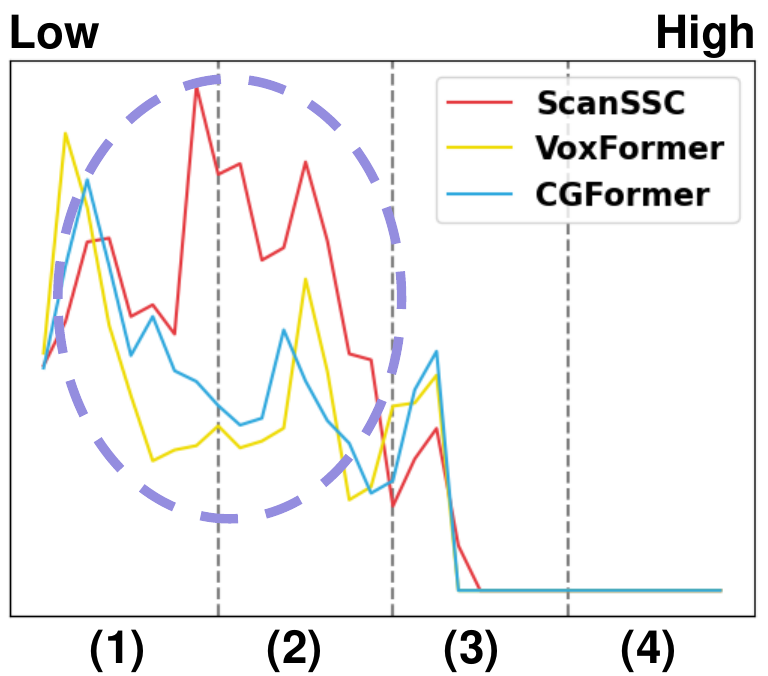}  }
        \label{fig:height_compared}
    \end{subfigure}
    
\vspace{-.3cm}
    \caption{Axis-wise trends in mIoU for VoxFormer~\cite{li2023voxformer}, CGFormer~\cite{yu2024contextgeometryawarevoxel}, and ScanSSC on the SemanticKITTI~\cite{behley2019semantickitti} validation data. Each graph is
binned with sizes of 256, 256, and 32 for the depth, width, and
height axes, respectively.}
    \label{fig:compared}
\vspace{-.4cm}
\end{figure}
\label{sec:results}
\begin{table}[hbt!]
\centering
\resizebox{\columnwidth}{!}{%
\begin{tabular}{c|ccc|ccc|ccc}
\hline
 &
  \multicolumn{3}{c|}{Depth} &
  \multicolumn{3}{c|}{Width} &
  \multicolumn{3}{c}{Height} \\ \cline{2-10} 
 &
  VoxFormer &
  CGFormer &
  ScanSSC &
  VoxFormer &
  CGFormer &
  ScanSSC &
  VoxFormer &
  CGFormer &
  ScanSSC \\ \hline
\large (1) &
   \large 12.84&
   \large 14.13&
   \large 13.77&
  \cellcolor[HTML]{F6EDF6}{\large 1.26} &
  \cellcolor[HTML]{F6EDF6}{\large 2.34} &
  \cellcolor[HTML]{F6EDF6}{\large \textbf{2.93}} &
   \cellcolor[HTML]{F6EDF6}{\large 9.55}&
   \cellcolor[HTML]{F6EDF6}{\large 10.85}&
   \cellcolor[HTML]{F6EDF6}{\large \textbf{12.30}}\\
\large (2) &
   \large 14.36&
   \large 13.59&
   \large 12.94&
   \large 5.04&
   \large 4.80&
   \large 5.84&
   \cellcolor[HTML]{F6EDF6}{\large 6.57}&
   \cellcolor[HTML]{F6EDF6}{\large 6.88}&
   \cellcolor[HTML]{F6EDF6}{\large \textbf{13.50}}\\
\large (3) &
   \cellcolor[HTML]{F6EDF6}{\large 12.27}&
   \cellcolor[HTML]{F6EDF6}{\large 12.67}&
   \cellcolor[HTML]{F6EDF6}{\large \textbf{13.50}}&
   \large 4.07&
   \large 6.72&
   \large 6.53&
   \large 2.87&
   \large 2.69&
   \large 2.07\\
\large (4) &
  \cellcolor[HTML]{F6EDF6}{\large 7.93} &
  \cellcolor[HTML]{F6EDF6}{\large 9.31} &
  \cellcolor[HTML]{F6EDF6}{\large \textbf{13.19}} &
  \cellcolor[HTML]{F6EDF6}{\large 0.65} &
  \cellcolor[HTML]{F6EDF6}{\large 3.42} &
  \cellcolor[HTML]{F6EDF6}{\large \textbf{4.80}} &
   \large 0.00&
   \large 0.00&
   \large 0.00\\ \hline
\end{tabular}%
}
\caption{For a comparison of mIoU values across each axis for VoxFormer~\cite{li2023voxformer}, CGFormer~\cite{yu2024contextgeometryawarevoxel}, and ScanSSC, we divided each axis into four segments labeled (1) through (4).}
\vspace{-0.2cm}
\label{tab:exp_distant}
\end{table}

We compare the performance of our ScanSSC with existing camera-based SSC methods~\cite{cao2022monoscene, huang2023tri, wei2023surroundocc, zhang2023occformer, xiao2024instance, li2023voxformer, wang2024not, zheng2024monoocc, wang2024h2gformer, jiang2024symphonize, li2024bridgingstereogeometrybev, yu2024contextgeometryawarevoxel, li2024hierarchical} on the SemanticKITTI~\cite{behley2019semantickitti} and SSCBench-KITTI-360 benchmarks~\cite{li2024sscbenchlargescale3dsemantic}. 
We list the results on the SemanticKITTI hidden test set in Tab.~\ref{tab:SemKITTI_test}.
ScanSSC achieves SOTA IoU and mIoU scores of 44.54 and 17.40, respectively, significantly outperforming all competing methods.
Among stereo-based (S) methods, ScanSSC stands out with an impressive performance improvement of 0.13 in IoU and 0.77 in mIoU, compared to CGFormer~\cite{yu2024contextgeometryawarevoxel}, the most competitive existing method.
Compared to HTCL-S~\cite{li2024hierarchical}, the SOTA temporal stereo-based (S-T) method, ScanSSC surpasses it even using only a single image input.
Furthermore, to evaluate the generalizability of ScanSSC, we also compare its performance on the SSCBench-KITTI-360 test dataset, as shown in Tab.~\ref{tab:KITTI360_test}. 
On this dataset, ScanSSC achieves SOTA IoU and mIoU scores of 48.29 and 20.14, respectively, notably surpassing other methods.
Additionally, to assess the robustness of ScanSSC in capturing distant geometric structures, we compare the axis-wise mIoU trend with other methods in Fig.~\ref{fig:compared}, while Tab.~\ref{tab:exp_distant} presents the averaged values across four segments (1)–(4) for each axis.
ScanSSC excels in achieving higher mIoU scores, especially in challenging distant regions.
These results provide quantitative evidence of the superiority of ScanSSC, especially in addressing the distance-dependent completion imbalance in camera-based SSC.

\subsection{Ablation Study}
\label{sec:ablation}
We conduct several ablation studies on ScanSSC using the SemanticKITTI validation set for all experiments.\\

\noindent\textbf{Ablation Study of Architectural Components.}
Tab.~\ref{tab:ablation_components} presents an ablation study on the architectural components, the Scan Module and Scan Loss.
Using the baseline model that excludes both components, we assess the impact of axis-wise subcomponents for the Scan Module ((a)-(c)) and Scan Loss ((e)-(g)).
Each subcomponent individually leads to a significant improvement in IoU and mIoU scores.
When evaluating the full Scan Module and Scan Loss, as shown in (d) and (h), the Scan Module contributes more to performance improvement, with IoU and mIoU increases of 3.86 and 1.72, respectively, compared to increases of 0.85 and 0.28 seen with Scan Loss.
Ultimately, by combining both components, a notable synergy is observed in the mIoU score, resulting in improvements of 3.72 in IoU and 2.21 in mIoU.
These results show that both components interact effectively for common objects, refining the semantics of distant voxels by leveraging the well-established context from near-viewpoint voxels.\\
\newcommand{\colorboxlabel}[2]{\raisebox{0pt}[0pt][0pt]{\tikz{\node[fill=#1, minimum width=0.28cm, minimum height=0.28cm, inner sep=0pt] {};}}~#2}
\newcommand{\rotatebigtext}[1]{\rotatebox{90}{\fontsize{8}{8}\selectfont #1}}
\newcommand{\rotatetinytext}[1]{\rotatebox{90}{\fontsize{4}{4}\selectfont #1}}
\newcommand{\rotateemptytext}[1]{\rotatebox{90}{\fontsize{2}{2}\selectfont #1}}
\newcommand{\rotatabovetext}[1]{\rotatebox{90}{\fontsize{5}{5}\selectfont #1}}

\definecolor{custom_road}{RGB}{255, 0, 255}
\definecolor{custom_sidewalk}{RGB}{75, 0, 75}
\definecolor{custom_parking}{RGB}{255, 150, 255}
\definecolor{custom_other_ground}{RGB}{175, 0, 75}
\definecolor{custom_building}{RGB}{255, 200, 0}
\definecolor{custom_car}{RGB}{100, 150, 245}
\definecolor{custom_truck}{RGB}{80, 30, 180}
\definecolor{custom_bicycle}{RGB}{100, 230, 245}
\definecolor{custom_motorcycle}{RGB}{30, 60, 150}
\definecolor{custom_other_veh}{RGB}{0, 0, 255}
\definecolor{custom_vegetation}{RGB}{0, 175, 0}
\definecolor{custom_trunk}{RGB}{135, 60, 0}
\definecolor{custom_terrain}{RGB}{150, 240, 80}
\definecolor{custom_person}{RGB}{255, 30, 30}
\definecolor{custom_bicyclist}{RGB}{255, 40, 20}
\definecolor{custom_motorcyclist}{RGB}{150, 30, 90}
\definecolor{custom_fence}{RGB}{255, 120, 50}
\definecolor{custom_pole}{RGB}{255, 240, 150}
\definecolor{custom_traf_sign}{RGB}{255, 0, 0}

\begin{table*}[hbt!]
\centering
\resizebox{\textwidth}{!}{%
\begin{tabular}{lccl|clclclclclclclclclclclclclclclclclclcl}
\hline
 &
&
  \multicolumn{1}{l}{} &
   &
  \multicolumn{2}{l}{} &
  \multicolumn{2}{l}{} &
  \multicolumn{2}{l}{} &
  \multicolumn{2}{l}{} &
  \multicolumn{2}{l}{} &
  \multicolumn{2}{l}{} &
  \multicolumn{2}{l}{} &
  \multicolumn{2}{l}{} &
  \multicolumn{2}{l}{} &
  \multicolumn{2}{l}{} &
  \multicolumn{2}{l}{} &
  \multicolumn{2}{l}{} &
  \multicolumn{2}{l}{} &
  \multicolumn{2}{l}{} &
  \multicolumn{2}{l}{} &
  \multicolumn{2}{l}{} &
  \multicolumn{2}{l}{} &
  \multicolumn{2}{l}{} &
  \multicolumn{2}{l}{} \\
 &
 &
  \multicolumn{1}{l}{} &
   &
  \multicolumn{2}{l}{\textcolor{white}{\rotatabovetext{1}} \newline \rotatebigtext{road} \newline \textcolor{white}{\rotatetinytext{1}} \newline \rotatetinytext{(15.30\%)}} &
  \multicolumn{2}{l}{\textcolor{white}{\rotatabovetext{1}} \newline \rotatebigtext{sidewalk} \newline \textcolor{white}{\rotatetinytext{1}} \newline \rotatetinytext{(11.13\%)}} &
  \multicolumn{2}{l}{\textcolor{white}{\rotatabovetext{1}} \newline \rotatebigtext{parking} \newline \textcolor{white}{\rotatetinytext{1}}\newline \rotatetinytext{(1.12\%)}} &
  \multicolumn{2}{l}{\textcolor{white}{\rotatabovetext{1}} \newline \rotatebigtext{other-grnd.} \newline \textcolor{white}{\rotatetinytext{1}}\newline \rotatetinytext{(0.56\%)}} &
  \multicolumn{2}{l}{\textcolor{white}{\rotatabovetext{1}} \newline \rotatebigtext{building} \newline \textcolor{white}{\rotatetinytext{1}}\newline \rotatetinytext{(14.1\%)}} &
  \multicolumn{2}{l}{\textcolor{white}{\rotatabovetext{1}} \newline \rotatebigtext{car} \newline \textcolor{white}{\rotatetinytext{1}}\newline \rotatetinytext{(3.92\%)}} &
  \multicolumn{2}{l}{\textcolor{white}{\rotatabovetext{1}} \newline \rotatebigtext{truck} \newline \textcolor{white}{\rotatetinytext{1}}\newline \rotatetinytext{(0.16\%)}} &
  \multicolumn{2}{l}{\textcolor{white}{\rotatabovetext{1}} \newline \rotatebigtext{bicycle} \newline \textcolor{white}{\rotatetinytext{1}}\newline \rotatetinytext{(0.03\%)}} &
  \multicolumn{2}{l}{\textcolor{white}{\rotatabovetext{1}} \newline \rotatebigtext{motorcycle} \newline \textcolor{white}{\rotatetinytext{1}}\newline \rotatetinytext{(0.03\%)}} &
  \multicolumn{2}{l}{\textcolor{white}{\rotatabovetext{1}} \newline \rotatebigtext{other-veh.} \newline \textcolor{white}{\rotatetinytext{1}}\newline \rotatetinytext{(0.20\%)}} &
  \multicolumn{2}{l}{\textcolor{white}{\rotatabovetext{1}} \newline \rotatebigtext{vegetation} \newline \textcolor{white}{\rotatetinytext{1}}\newline \rotatetinytext{(39.3\%)}} &
  \multicolumn{2}{l}{\textcolor{white}{\rotatabovetext{1}} \newline \rotatebigtext{trunk} \newline \textcolor{white}{\rotatetinytext{1}}\newline \rotatetinytext{(0.51\%)}} &
  \multicolumn{2}{l}{\textcolor{white}{\rotatabovetext{1}} \newline \rotatebigtext{terrain} \newline \textcolor{white}{\rotatetinytext{1}}\newline \rotatetinytext{(9.17\%)}} &
  \multicolumn{2}{l}{\textcolor{white}{\rotatabovetext{1}} \newline \rotatebigtext{person} \newline \textcolor{white}{\rotatetinytext{1}}\newline \rotatetinytext{(0.07\%)}} &
  \multicolumn{2}{l}{\textcolor{white}{\rotatabovetext{1}} \newline \rotatebigtext{bicyclist} \newline \textcolor{white}{\rotatetinytext{1}}\newline \rotatetinytext{(0.07\%)}} &
  \multicolumn{2}{l}{\textcolor{white}{\rotatabovetext{1}} \newline \rotatebigtext{motorcyclist} \newline \textcolor{white}{\rotatetinytext{1}}\newline \rotatetinytext{(0.05\%)}} &
  \multicolumn{2}{l}{\textcolor{white}{\rotatabovetext{1}} \newline \rotatebigtext{fence} \newline \textcolor{white}{\rotatetinytext{1}}\newline \rotatetinytext{(3.90\%)}} &
  \multicolumn{2}{l}{\textcolor{white}{\rotatabovetext{1}} \newline \rotatebigtext{pole} \newline \textcolor{white}{\rotatetinytext{1}}\newline \rotatetinytext{(0.29\%)}} &
  \multicolumn{2}{l}{\textcolor{white}{\rotatabovetext{1}} \newline \rotatebigtext{traf.-sign} \newline \textcolor{white}{\rotatetinytext{1}}\newline \rotatetinytext{(0.08\%)}} \\
Method &
Input&
  IoU &
  \multicolumn{1}{c|}{mIoU} &
  \multicolumn{2}{c}{\colorboxlabel{custom_road}{}} &
  \multicolumn{2}{c}{\colorboxlabel{custom_sidewalk}{}} &
  \multicolumn{2}{c}{\colorboxlabel{custom_parking}{}} &
  \multicolumn{2}{c}{\colorboxlabel{custom_other_ground}{}} &
  \multicolumn{2}{c}{\colorboxlabel{custom_building}{}} &
  \multicolumn{2}{c}{\colorboxlabel{custom_car}{}} &
  \multicolumn{2}{c}{\colorboxlabel{custom_truck}{}} &
  \multicolumn{2}{c}{\colorboxlabel{custom_bicycle}{}} &
  \multicolumn{2}{c}{\colorboxlabel{custom_motorcycle}{}} &
  \multicolumn{2}{c}{\colorboxlabel{custom_other_veh}{}} &
  \multicolumn{2}{c}{\colorboxlabel{custom_vegetation}{}} &
  \multicolumn{2}{c}{\colorboxlabel{custom_trunk}{}} &
  \multicolumn{2}{c}{\colorboxlabel{custom_terrain}{}} &
  \multicolumn{2}{c}{\colorboxlabel{custom_person}{}} &
  \multicolumn{2}{c}{\colorboxlabel{custom_bicyclist}{}} &
  \multicolumn{2}{c}{\colorboxlabel{custom_motorcyclist}{}} &
  \multicolumn{2}{c}{\colorboxlabel{custom_fence}{}} &
  \multicolumn{2}{c}{\colorboxlabel{custom_pole}{}} &
  \multicolumn{2}{c}{\colorboxlabel{custom_traf_sign}{}} \\ \hline

 \multicolumn{1}{l|}{MonoScene~\cite{cao2022monoscene}} &
 \multicolumn{1}{c|}{M}&
  34.16 &
  \multicolumn{1}{c|}{11.08} &
  \multicolumn{2}{c}{54.70} &
  \multicolumn{2}{c}{27.10} &
  \multicolumn{2}{c}{24.80} &
  \multicolumn{2}{c}{5.70} &
  \multicolumn{2}{c}{14.40} &
  \multicolumn{2}{c}{18.80} &
  \multicolumn{2}{c}{3.30} &
  \multicolumn{2}{c}{0.50} &
  \multicolumn{2}{c}{0.70} &
  \multicolumn{2}{c}{4.40} &
  \multicolumn{2}{c}{14.90} &
  \multicolumn{2}{c}{2.40} &
  \multicolumn{2}{c}{19.50} &
  \multicolumn{2}{c}{1.00} &
  \multicolumn{2}{c}{1.40} &
  \multicolumn{2}{c}{0.40} &
  \multicolumn{2}{c}{11.10} &
  \multicolumn{2}{c}{3.30} &
  \multicolumn{2}{c}{2.10} \\
 \multicolumn{1}{l|}{TPVFormer~\cite{huang2023tri}} &
 \multicolumn{1}{c|}{M}&
  34.25 &
  \multicolumn{1}{c|}{11.26} &
  \multicolumn{2}{c}{55.10} &
  \multicolumn{2}{c}{27.20} &
  \multicolumn{2}{c}{27.40} &
  \multicolumn{2}{c}{6.50} &
  \multicolumn{2}{c}{14.80} &
  \multicolumn{2}{c}{19.20} &
  \multicolumn{2}{c}{3.70} &
  \multicolumn{2}{c}{1.00} &
  \multicolumn{2}{c}{0.50} &
  \multicolumn{2}{c}{2.30} &
  \multicolumn{2}{c}{13.90} &
  \multicolumn{2}{c}{2.60} &
  \multicolumn{2}{c}{20.40} &
  \multicolumn{2}{c}{1.10} &
  \multicolumn{2}{c}{2.40} &
  \multicolumn{2}{c}{0.30} &
  \multicolumn{2}{c}{11.00} &
  \multicolumn{2}{c}{2.90} &
  \multicolumn{2}{c}{1.50} \\
 \multicolumn{1}{l|}{SurroundOcc~\cite{wei2023surroundocc}} &
 \multicolumn{1}{c|}{M}&
  34.72 &
  \multicolumn{1}{c|}{11.86} &
  \multicolumn{2}{c}{56.90} &
  \multicolumn{2}{c}{28.30} &
  \multicolumn{2}{c}{30.20} &
  \multicolumn{2}{c}{6.80} &
  \multicolumn{2}{c}{15.20} &
  \multicolumn{2}{c}{20.60} &
  \multicolumn{2}{c}{1.40} &
  \multicolumn{2}{c}{1.60} &
  \multicolumn{2}{c}{1.20} &
  \multicolumn{2}{c}{4.40} &
  \multicolumn{2}{c}{14.90} &
  \multicolumn{2}{c}{3.40} &
  \multicolumn{2}{c}{19.30} &
  \multicolumn{2}{c}{1.40} &
  \multicolumn{2}{c}{2.00} &
  \multicolumn{2}{c}{0.10} &
  \multicolumn{2}{c}{11.30} &
  \multicolumn{2}{c}{3.90} &
  \multicolumn{2}{c}{2.40} \\
\multicolumn{1}{l|}{OccFormer~\cite{zhang2023occformer}} &
\multicolumn{1}{c|}{M}&
  34.53 &
  \multicolumn{1}{c|}{12.32} &
  \multicolumn{2}{c}{55.90} &
  \multicolumn{2}{c}{30.30} &
  \multicolumn{2}{c}{31.50} &
  \multicolumn{2}{c}{6.50} &
  \multicolumn{2}{c}{15.70} &
  \multicolumn{2}{c}{21.60} &
  \multicolumn{2}{c}{1.20} &
  \multicolumn{2}{c}{1.50} &
  \multicolumn{2}{c}{1.70} &
  \multicolumn{2}{c}{3.20} &
  \multicolumn{2}{c}{16.80} &
  \multicolumn{2}{c}{3.90} &
  \multicolumn{2}{c}{21.30} &
  \multicolumn{2}{c}{2.20} &
  \multicolumn{2}{c}{1.10} &
  \multicolumn{2}{c}{0.20} &
  \multicolumn{2}{c}{11.90} &
  \multicolumn{2}{c}{3.80} &
  \multicolumn{2}{c}{3.70} \\
\multicolumn{1}{l|}{IAMSSC~\cite{xiao2024instance}} &
\multicolumn{1}{c|}{M}&
  43.74 &
  \multicolumn{1}{c|}{12.37} &
  \multicolumn{2}{c}{54.00} &
  \multicolumn{2}{c}{25.50} &
  \multicolumn{2}{c}{24.70} &
  \multicolumn{2}{c}{6.90} &
  \multicolumn{2}{c}{19.20} &
  \multicolumn{2}{c}{21.30} &
  \multicolumn{2}{c}{3.80} &
  \multicolumn{2}{c}{1.10} &
  \multicolumn{2}{c}{0.60} &
  \multicolumn{2}{c}{3.90} &
  \multicolumn{2}{c}{22.70} &
  \multicolumn{2}{c}{5.80} &
  \multicolumn{2}{c}{19.40} &
  \multicolumn{2}{c}{1.50} &
  \multicolumn{2}{c}{2.90} &
  \multicolumn{2}{c}{0.50} &
  \multicolumn{2}{c}{11.90} &
  \multicolumn{2}{c}{5.30} &
  \multicolumn{2}{c}{4.10} \\
\multicolumn{1}{l|}{VoxFormer-T~\cite{li2023voxformer}} &
\multicolumn{1}{c|}{S-T}&
  43.21 &
  \multicolumn{1}{c|}{13.41} &
  \multicolumn{2}{c}{54.10} &
  \multicolumn{2}{c}{26.90} &
  \multicolumn{2}{c}{25.10} &
  \multicolumn{2}{c}{7.30} &
  \multicolumn{2}{c}{23.50} &
  \multicolumn{2}{c}{21.70} &
  \multicolumn{2}{c}{3.60} &
  \multicolumn{2}{c}{1.90} &
  \multicolumn{2}{c}{1.60} &
  \multicolumn{2}{c}{4.10} &
  \multicolumn{2}{c}{24.40} &
  \multicolumn{2}{c}{8.10} &
  \multicolumn{2}{c}{24.20} &
  \multicolumn{2}{c}{1.60} &
  \multicolumn{2}{c}{1.10} &
  \multicolumn{2}{c}{0.00} &
  \multicolumn{2}{c}{13.10} &
  \multicolumn{2}{c}{6.60} &
  \multicolumn{2}{c}{5.70} \\
 \multicolumn{1}{l|}{HASSC-T~\cite{wang2024not}} &
 \multicolumn{1}{c|}{S-T}&
  42.87 &
  \multicolumn{1}{c|}{14.38} &
  \multicolumn{2}{c}{55.30} &
  \multicolumn{2}{c}{29.60} &
  \multicolumn{2}{c}{25.90} &
  \multicolumn{2}{c}{11.30} &
  \multicolumn{2}{c}{23.10} &
  \multicolumn{2}{c}{23.00} &
  \multicolumn{2}{c}{2.90} &
  \multicolumn{2}{c}{1.90} &
  \multicolumn{2}{c}{1.50} &
  \multicolumn{2}{c}{4.90} &
  \multicolumn{2}{c}{24.80} &
  \multicolumn{2}{c}{9.80} &
  \multicolumn{2}{c}{26.50} &
  \multicolumn{2}{c}{1.40} &
  \multicolumn{2}{c}{3.00} &
  \multicolumn{2}{c}{0.00} &
  \multicolumn{2}{c}{14.30} &
  \multicolumn{2}{c}{7.00} &
  \multicolumn{2}{c}{7.10} \\
 
 \multicolumn{1}{l|}{H2GFormer-T~\cite{wang2024h2gformer}} &
 \multicolumn{1}{c|}{S-T}&
  43.52 &
  \multicolumn{1}{c|}{14.60} &
  \multicolumn{2}{c}{57.90} &
  \multicolumn{2}{c}{30.40} &
  \multicolumn{2}{c}{30.00} &
  \multicolumn{2}{c}{6.90} &
  \multicolumn{2}{c}{24.00} &
  \multicolumn{2}{c}{23.70} &
  \multicolumn{2}{c}{5.20} &
  \multicolumn{2}{c}{0.60} &
  \multicolumn{2}{c}{1.20} &
  \multicolumn{2}{c}{5.00} &
  \multicolumn{2}{c}{\underline{25.20}} &
  \multicolumn{2}{c}{10.70} &
  \multicolumn{2}{c}{25.80} &
  \multicolumn{2}{c}{1.10} &
  \multicolumn{2}{c}{0.10} &
  \multicolumn{2}{c}{0.00} &
  \multicolumn{2}{c}{14.60} &
  \multicolumn{2}{c}{7.50} &
  \multicolumn{2}{c}{\textbf{{9.30}}} \\
 \multicolumn{1}{l|}{Symphonies~\cite{jiang2024symphonize}} &
 \multicolumn{1}{c|}{S}&
  42.19 &
  \multicolumn{1}{c|}{15.04} &
  \multicolumn{2}{c}{58.40} &
  \multicolumn{2}{c}{29.30} &
  \multicolumn{2}{c}{26.90} &
  \multicolumn{2}{c}{11.70} &
  \multicolumn{2}{c}{24.70} &
  \multicolumn{2}{c}{23.60} &
  \multicolumn{2}{c}{3.20} &
  \multicolumn{2}{c}{3.60} &
  \multicolumn{2}{c}{\underline{{2.60}}} &
  \multicolumn{2}{c}{5.60} &
  \multicolumn{2}{c}{24.20} &
  \multicolumn{2}{c}{10.00} &
  \multicolumn{2}{c}{23.10} &
  \multicolumn{2}{c}{\textbf{{3.20}}} &
  \multicolumn{2}{c}{1.90} &
  \multicolumn{2}{c}{\textbf{{2.00}}} &
  \multicolumn{2}{c}{16.10} &
  \multicolumn{2}{c}{7.70} &
  \multicolumn{2}{c}{8.00} \\
 \multicolumn{1}{l|}{StereoScene~\cite{li2024bridgingstereogeometrybev}} &
 \multicolumn{1}{c|}{S}&
  43.34 &
  \multicolumn{1}{c|}{15.36} &
  \multicolumn{2}{c}{61.90} &
  \multicolumn{2}{c}{31.20} &
  \multicolumn{2}{c}{30.70} &
  \multicolumn{2}{c}{10.70} &
  \multicolumn{2}{c}{24.20} &
  \multicolumn{2}{c}{22.80} &
  \multicolumn{2}{c}{2.80} &
  \multicolumn{2}{c}{3.40} &
  \multicolumn{2}{c}{2.40} &
  \multicolumn{2}{c}{\underline{{6.10}}} &
  \multicolumn{2}{c}{23.80} &
  \multicolumn{2}{c}{8.40} &
  \multicolumn{2}{c}{27.00} &
  \multicolumn{2}{c}{\underline{{2.90}}} &
  \multicolumn{2}{c}{2.20} &
  \multicolumn{2}{c}{0.50} &
  \multicolumn{2}{c}{16.50} &
  \multicolumn{2}{c}{7.00} &
  \multicolumn{2}{c}{7.20} \\
\multicolumn{1}{l|}{MonoOcc-L~\cite{zheng2024monoocc}} &
 \multicolumn{1}{c|}{S}&
  - &
  \multicolumn{1}{c|}{15.63} &
  \multicolumn{2}{c}{59.10} &
  \multicolumn{2}{c}{30.90} &
  \multicolumn{2}{c}{27.10} &
  \multicolumn{2}{c}{9.80} &
  \multicolumn{2}{c}{22.90} &
  \multicolumn{2}{c}{23.90} &
  \multicolumn{2}{c}{\underline{{7.20}}} &
  \multicolumn{2}{c}{\textbf{{4.50}}} &
  \multicolumn{2}{c}{2.40} &
  \multicolumn{2}{c}{\textbf{{7.70}}} &
  \multicolumn{2}{c}{25.00} &
  \multicolumn{2}{c}{9.80} &
  \multicolumn{2}{c}{26.10} &
  \multicolumn{2}{c}{2.80} &
  \multicolumn{2}{c}{\underline{{4.70}}} &
  \multicolumn{2}{c}{0.60} &
  \multicolumn{2}{c}{16.90} &
  \multicolumn{2}{c}{7.30} &
  \multicolumn{2}{c}{8.40} \\
 \multicolumn{1}{l|}{CGFormer~\cite{yu2024contextgeometryawarevoxel}} &
 \multicolumn{1}{c|}{S}&
  \underline{{44.41}} &
  \multicolumn{1}{c|}{16.63} &
  \multicolumn{2}{c}{64.30} &
  \multicolumn{2}{c}{34.20} &
  \multicolumn{2}{c}{\underline{{34.10}}} &
  \multicolumn{2}{c}{12.10} &
  \multicolumn{2}{c}{\underline{{25.80}}} &
  \multicolumn{2}{c}{26.10} &
  \multicolumn{2}{c}{4.30} &
  \multicolumn{2}{c}{\underline{{3.70}}} &
  \multicolumn{2}{c}{1.30} &
  \multicolumn{2}{c}{2.70} &
  \multicolumn{2}{c}{24.50} &
  \multicolumn{2}{c}{\textbf{{11.20}}} &
  \multicolumn{2}{c}{29.30} &
  \multicolumn{2}{c}{1.70} &
  \multicolumn{2}{c}{3.60} &
  \multicolumn{2}{c}{0.40} &
  \multicolumn{2}{c}{18.70} &
  \multicolumn{2}{c}{\underline{{8.70}}} &
  \multicolumn{2}{c}{\textbf{{9.30}}} \\
\multicolumn{1}{l|}{HTCL-S~\cite{li2024hierarchical}} &
\multicolumn{1}{c|}{S-T}&
  44.23 &
  \multicolumn{1}{c|}{\underline{{17.09}}} &
  \multicolumn{2}{c}{\underline{{64.40}}} &
  \multicolumn{2}{c}{\underline{{34.80}}} &
  \multicolumn{2}{c}{33.80} &
  \multicolumn{2}{c}{\underline{{12.40}}} &
  \multicolumn{2}{c}{\textbf{{25.90}}} &
  \multicolumn{2}{c}{\textbf{{27.30}}} &
  \multicolumn{2}{c}{\textbf{{10.80}}} &
  \multicolumn{2}{c}{1.80} &
  \multicolumn{2}{c}{2.20} &
  \multicolumn{2}{c}{5.40} &
  \multicolumn{2}{c}{\textbf{{25.30}}} &
  \multicolumn{2}{c}{10.80} &
  \multicolumn{2}{c}{\textbf{{31.20}}} &
  \multicolumn{2}{c}{1.10} &
  \multicolumn{2}{c}{3.10} &
  \multicolumn{2}{c}{\underline{{0.90}}} &
  \multicolumn{2}{c}{\textbf{{21.10}}} &
  \multicolumn{2}{c}{\textbf{{9.00}}} &
  \multicolumn{2}{c}{8.30} \\

 \rowcolor{violet!7}\multicolumn{1}{l|}{\textbf{ScanSSC (ours)}}&
 \multicolumn{1}{c|}{S}&
 \textbf{{44.54}} &
  \multicolumn{1}{c|}{\textbf{{17.40}}} &
  \multicolumn{2}{c}{\textbf{{66.20}}} &
  \multicolumn{2}{c}{\textbf{{35.90}}} &
  \multicolumn{2}{c}{\textbf{{35.10}}} &
  \multicolumn{2}{c}{\textbf{{12.50}}} &
  \multicolumn{2}{c}{25.30} &
  \multicolumn{2}{c}{\underline{{27.10}}} &
  \multicolumn{2}{c}{3.50} &
  \multicolumn{2}{c}{3.50} &
  \multicolumn{2}{c}{\textbf{{3.20}}} &
  \multicolumn{2}{c}{\underline{{6.10}}} &
  \multicolumn{2}{c}{\underline{{25.20}}} &
  \multicolumn{2}{c}{\underline{{11.00}}} &
  \multicolumn{2}{c}{\underline{{30.60}}} &
  \multicolumn{2}{c}{1.80} &
  \multicolumn{2}{c}{\textbf{{5.30}}} &
  \multicolumn{2}{c}{0.70} &
  \multicolumn{2}{c}{\underline{{20.50}}} &
  \multicolumn{2}{c}{8.40} &
  \multicolumn{2}{c}{\underline{{8.90}}} \\ \hline
\end{tabular}%
}
\vspace{-0.2cm}
\caption{Quantitative results on SemanticKITTI hidden test set. `M', `S', and `S-T' represent the monocular, stereo, and temporal stereo inputs, respectively. \textbf{Bold} and \underline{underline} highlight the best and second-best results, respectively.}
\vspace{-0.2cm}
\label{tab:SemKITTI_test}
    
\end{table*}


\definecolor{custom_road}{RGB}{255, 0, 255}
\definecolor{custom_sidewalk}{RGB}{75, 0, 75}
\definecolor{custom_parking}{RGB}{255, 150, 255}
\definecolor{custom_other_ground}{RGB}{175, 0, 75}
\definecolor{custom_building}{RGB}{255, 200, 0}
\definecolor{custom_car}{RGB}{100, 150, 245}
\definecolor{custom_truck}{RGB}{80, 30, 180}
\definecolor{custom_bicycle}{RGB}{100, 230, 245}
\definecolor{custom_motorcycle}{RGB}{30, 60, 150}
\definecolor{custom_other_veh}{RGB}{0, 0, 255}
\definecolor{custom_vegetation}{RGB}{0, 175, 0}
\definecolor{custom_trunk}{RGB}{135, 60, 0}
\definecolor{custom_terrain}{RGB}{150, 240, 80}
\definecolor{custom_person}{RGB}{255, 30, 30}
\definecolor{custom_bicyclist}{RGB}{255, 40, 20}
\definecolor{custom_motorcyclist}{RGB}{150, 30, 90}
\definecolor{custom_fence}{RGB}{255, 120, 50}
\definecolor{custom_pole}{RGB}{255, 240, 150}
\definecolor{custom_traf_sign}{RGB}{255, 0, 0}
\definecolor{custom_other_struct}{RGB}{255, 150, 0}
\definecolor{custom_other_obj}{RGB}{50, 255, 255}

\begin{table*}[t]
\centering
\resizebox{\textwidth}{!}{%
\begin{tabular}{lccc|cccccccccccccccccccccccccccccccccccccc}
\hline
 &
 &
  \multicolumn{1}{l}{} &
   &
  \multicolumn{2}{c}{} &
  \multicolumn{2}{c}{} &
  \multicolumn{2}{c}{} &
  \multicolumn{2}{c}{} &
  \multicolumn{2}{c}{} &
  \multicolumn{2}{c}{} &
  \multicolumn{2}{c}{} &
  \multicolumn{2}{c}{} &
  \multicolumn{2}{c}{} &
  \multicolumn{2}{c}{} &
  \multicolumn{2}{c}{} &
  \multicolumn{2}{c}{} &
  \multicolumn{2}{c}{} &
  \multicolumn{2}{c}{} &
  \multicolumn{2}{c}{} &
  \multicolumn{2}{c}{} &
  \multicolumn{2}{c}{} &
  \multicolumn{2}{c}{} &
  \multicolumn{2}{c}{} \\
 &
 &
  \multicolumn{1}{c}{} &
   &
  \multicolumn{2}{c}{\rotatebigtext{car} \newline \textcolor{white}{\rotatetinytext{1}} \newline \rotatetinytext{(2.85\%)}} &
  \multicolumn{2}{c}{\rotatebigtext{bicycle} \newline \textcolor{white}{\rotatetinytext{1}} \newline \rotatetinytext{(0.01\%)}} &
  \multicolumn{2}{c}{\rotatebigtext{motorcycle} \newline \textcolor{white}{\rotatetinytext{1}}\newline \rotatetinytext{(0.01\%)}} &
  \multicolumn{2}{c}{\rotatebigtext{truck} \newline \textcolor{white}{\rotatetinytext{1}}\newline \rotatetinytext{(0.16\%)}} &
  \multicolumn{2}{c}{\rotatebigtext{other-veh.} \newline \textcolor{white}{\rotatetinytext{1}}\newline \rotatetinytext{(5.75\%)}} &
  \multicolumn{2}{c}{\rotatebigtext{person} \newline \textcolor{white}{\rotatetinytext{1}}\newline \rotatetinytext{(0.02\%)}} &
  \multicolumn{2}{c}{\rotatebigtext{road} \newline \textcolor{white}{\rotatetinytext{1}}\newline \rotatetinytext{(14.98\%)}} &
  \multicolumn{2}{c}{\rotatebigtext{parking} \newline \textcolor{white}{\rotatetinytext{1}}\newline \rotatetinytext{(2.31\%)}} &
  \multicolumn{2}{c}{\rotatebigtext{sidewalk} \newline \textcolor{white}{\rotatetinytext{1}}\newline \rotatetinytext{(6.43\%)}} &
  \multicolumn{2}{c}{\rotatebigtext{other-grnd.} \newline \textcolor{white}{\rotatetinytext{1}}\newline \rotatetinytext{(2.05\%)}} &
  \multicolumn{2}{c}{\rotatebigtext{building} \newline \textcolor{white}{\rotatetinytext{1}}\newline \rotatetinytext{(15.67\%)}} &
  \multicolumn{2}{c}{\rotatebigtext{fence} \newline \textcolor{white}{\rotatetinytext{1}}\newline \rotatetinytext{(0.96\%)}} &
  \multicolumn{2}{c}{\rotatebigtext{vegetation} \newline \textcolor{white}{\rotatetinytext{1}}\newline \rotatetinytext{(41.99\%)}} &
  \multicolumn{2}{c}{\rotatebigtext{terrain} \newline \textcolor{white}{\rotatetinytext{1}}\newline \rotatetinytext{(7.10\%)}} &
  \multicolumn{2}{c}{\rotatebigtext{pole} \newline \textcolor{white}{\rotatetinytext{1}}\newline \rotatetinytext{(0.22\%)}} &
  \multicolumn{2}{c}{\rotatebigtext{traf.-sign} \newline \textcolor{white}{\rotatetinytext{1}}\newline \rotatetinytext{(0.06\%)}} &
  \multicolumn{2}{c}{\rotatebigtext{other-struct.} \newline \textcolor{white}{\rotatetinytext{1}}\newline \rotatetinytext{(4.33\%)}} &
  \multicolumn{2}{c}{\rotatebigtext{other-obj.} \newline \textcolor{white}{\rotatetinytext{1}}\newline \rotatetinytext{(0.28\%)}} &
\\
Method &
Input&
  IoU &
  \multicolumn{1}{c|}{mIoU} &
  \multicolumn{2}{c}{\colorboxlabel{custom_car}{}} &
  \multicolumn{2}{c}{\colorboxlabel{custom_bicycle}{}} &
  \multicolumn{2}{c}{\colorboxlabel{custom_motorcycle}{}} &
  \multicolumn{2}{c}{\colorboxlabel{custom_truck}{}} &
  \multicolumn{2}{c}{\colorboxlabel{custom_other_veh}{}} &
  \multicolumn{2}{c}{\colorboxlabel{custom_person}{}} &
  \multicolumn{2}{c}{\colorboxlabel{custom_road}{}} &
  \multicolumn{2}{c}{\colorboxlabel{custom_parking}{}} &
  \multicolumn{2}{c}{\colorboxlabel{custom_sidewalk}{}} &
  \multicolumn{2}{c}{\colorboxlabel{custom_other_ground}{}} &
  \multicolumn{2}{c}{\colorboxlabel{custom_building}{}} &
  \multicolumn{2}{c}{\colorboxlabel{custom_fence}{}} &
  \multicolumn{2}{c}{\colorboxlabel{custom_vegetation}{}} &
  \multicolumn{2}{c}{\colorboxlabel{custom_terrain}{}} &
  \multicolumn{2}{c}{\colorboxlabel{custom_pole}{}} &
  \multicolumn{2}{c}{\colorboxlabel{custom_traf_sign}{}} &
  \multicolumn{2}{c}{\colorboxlabel{custom_other_struct}{}} &
  \multicolumn{2}{c}{\colorboxlabel{custom_other_obj}{}} &
 \\ \hline
 \multicolumn{1}{l|}{MonoScene~\cite{cao2022monoscene}} &
 \multicolumn{1}{c|}{M}&
  37.87 &
  \multicolumn{1}{c|}{12.31} &
  \multicolumn{2}{c}{19.34} &
  \multicolumn{2}{c}{0.43} &
  \multicolumn{2}{c}{0.58} &
  \multicolumn{2}{c}{8.02} &
  \multicolumn{2}{c}{2.03} &
  \multicolumn{2}{c}{0.86} &
  \multicolumn{2}{c}{48.35} &
  \multicolumn{2}{c}{11.38} &
  \multicolumn{2}{c}{28.13} &
  \multicolumn{2}{c}{3.32} &
  \multicolumn{2}{c}{32.89} &
  \multicolumn{2}{c}{3.53} &
  \multicolumn{2}{c}{26.15} &
  \multicolumn{2}{c}{16.75} &
  \multicolumn{2}{c}{6.92} &
  \multicolumn{2}{c}{5.67} &
  \multicolumn{2}{c}{4.20} &
  \multicolumn{2}{c}{3.09} &
\\
 \multicolumn{1}{l|}{TPVFormer~\cite{huang2023tri}} &
 \multicolumn{1}{c|}{M}&
  40.22 &
  \multicolumn{1}{c|}{13.64} &
  \multicolumn{2}{c}{21.56} &
  \multicolumn{2}{c}{1.09} &
  \multicolumn{2}{c}{1.37} &
  \multicolumn{2}{c}{8.06} &
  \multicolumn{2}{c}{2.57} &
  \multicolumn{2}{c}{2.38} &
  \multicolumn{2}{c}{52.99} &
  \multicolumn{2}{c}{11.99} &
  \multicolumn{2}{c}{31.07} &
  \multicolumn{2}{c}{3.78} &
  \multicolumn{2}{c}{34.83} &
  \multicolumn{2}{c}{4.80} &
  \multicolumn{2}{c}{30.08} &
  \multicolumn{2}{c}{17.52} &
  \multicolumn{2}{c}{7.46} &
  \multicolumn{2}{c}{5.86} &
  \multicolumn{2}{c}{5.48} &
  \multicolumn{2}{c}{2.70} &
 \\
\multicolumn{1}{l|}{OccFormer~\cite{zhang2023occformer}} &
\multicolumn{1}{c|}{M}&
  40.27 &
  \multicolumn{1}{c|}{13.81} &
  \multicolumn{2}{c}{22.58} &
  \multicolumn{2}{c}{0.66} &
  \multicolumn{2}{c}{0.26} &
  \multicolumn{2}{c}{9.89} &
  \multicolumn{2}{c}{3.82} &
  \multicolumn{2}{c}{2.77} &
  \multicolumn{2}{c}{54.30} &
  \multicolumn{2}{c}{13.44} &
  \multicolumn{2}{c}{31.53} &
  \multicolumn{2}{c}{3.55} &
  \multicolumn{2}{c}{36.42} &
  \multicolumn{2}{c}{4.80} &
  \multicolumn{2}{c}{31.00} &
  \multicolumn{2}{c}{19.51} &
  \multicolumn{2}{c}{7.77} &
  \multicolumn{2}{c}{8.51} &
  \multicolumn{2}{c}{6.95} &
  \multicolumn{2}{c}{4.60} &
 \\
\multicolumn{1}{l|}{VoxFormer~\cite{li2023voxformer}} &
\multicolumn{1}{c|}{S}&
  38.76 &
  \multicolumn{1}{c|}{11.91} &
  \multicolumn{2}{c}{17.84} &
  \multicolumn{2}{c}{1.16} &
  \multicolumn{2}{c}{0.89} &
  \multicolumn{2}{c}{4.56} &
  \multicolumn{2}{c}{2.06} &
  \multicolumn{2}{c}{1.63} &
  \multicolumn{2}{c}{47.01} &
  \multicolumn{2}{c}{9.67} &
  \multicolumn{2}{c}{27.21} &
  \multicolumn{2}{c}{2.89} &
  \multicolumn{2}{c}{31.18} &
  \multicolumn{2}{c}{4.97} &
  \multicolumn{2}{c}{28.99} &
  \multicolumn{2}{c}{14.69} &
  \multicolumn{2}{c}{6.51} &
  \multicolumn{2}{c}{6.92} &
  \multicolumn{2}{c}{3.79} &
  \multicolumn{2}{c}{2.43} &
 \\
 \multicolumn{1}{l|}{IAMSSC~\cite{xiao2024instance}} &
 \multicolumn{1}{c|}{M}&
  41.80 &
  \multicolumn{1}{c|}{12.97} &
  \multicolumn{2}{c}{18.53} &
  \multicolumn{2}{c}{2.45} &
  \multicolumn{2}{c}{1.76} &
  \multicolumn{2}{c}{5.12} &
  \multicolumn{2}{c}{3.92} &
  \multicolumn{2}{c}{3.09} &
  \multicolumn{2}{c}{47.55} &
  \multicolumn{2}{c}{10.56} &
  \multicolumn{2}{c}{28.35} &
  \multicolumn{2}{c}{4.12} &
  \multicolumn{2}{c}{31.53} &
  \multicolumn{2}{c}{6.28} &
  \multicolumn{2}{c}{29.17} &
  \multicolumn{2}{c}{15.24} &
  \multicolumn{2}{c}{8.29} &
  \multicolumn{2}{c}{7.01} &
  \multicolumn{2}{c}{6.35} &
  \multicolumn{2}{c}{4.19} &
\\
 \multicolumn{1}{l|}{Symphonize~\cite{jiang2024symphonize}} &
 \multicolumn{1}{c|}{S}&
  44.12 &
  \multicolumn{1}{c|}{18.58} &
  \multicolumn{2}{c}{\textbf{{30.02}}} &
  \multicolumn{2}{c}{1.85} &
  \multicolumn{2}{c}{\textbf{{5.90}}} &
  \multicolumn{2}{c}{\textbf{{25.07}}} &
  \multicolumn{2}{c}{\textbf{{12.06}}} &
  \multicolumn{2}{c}{\textbf{{8.20}}} &
  \multicolumn{2}{c}{54.94} &
  \multicolumn{2}{c}{13.83} &
  \multicolumn{2}{c}{32.76} &
  \multicolumn{2}{c}{\textbf{{6.93}}} &
  \multicolumn{2}{c}{35.11} &
  \multicolumn{2}{c}{\underline{{8.58}}} &
  \multicolumn{2}{c}{38.33} &
  \multicolumn{2}{c}{11.52} &
  \multicolumn{2}{c}{14.01} &
  \multicolumn{2}{c}{9.57} &
  \multicolumn{2}{c}{\textbf{{14.44}}} &
  \multicolumn{2}{c}{\textbf{{11.28}}} &
\\
 \multicolumn{1}{l|}{CGFormer~\cite{yu2024contextgeometryawarevoxel}} &
 \multicolumn{1}{c|}{S}&
  \underline{{48.07}} &
  \multicolumn{1}{c|}{\underline{{20.05}}} &
  \multicolumn{2}{c}{29.85} &
  \multicolumn{2}{c}{\underline{{3.42}}} &
  \multicolumn{2}{c}{3.96} &
  \multicolumn{2}{c}{\underline{{17.59}}} &
  \multicolumn{2}{c}{6.79} &
  \multicolumn{2}{c}{6.63} &
  \multicolumn{2}{c}{\textbf{{63.85}}} &
  \multicolumn{2}{c}{\underline{{17.15}}} &
  \multicolumn{2}{c}{\textbf{{40.72}}} &
  \multicolumn{2}{c}{\underline{{5.53}}} &
  \multicolumn{2}{c}{\textbf{{42.73}}} &
  \multicolumn{2}{c}{8.22} &
  \multicolumn{2}{c}{\underline{{38.80}}} &
  \multicolumn{2}{c}{\underline{{24.94}}} &
  \multicolumn{2}{c}{\underline{{16.24}}} &
  \multicolumn{2}{c}{\underline{{17.45}}} &
  \multicolumn{2}{c}{10.18} &
  \multicolumn{2}{c}{6.77} &
 \\ 
 \rowcolor{violet!7}\multicolumn{1}{l|}{\textbf{ScanSSC (ours)}}&
 \multicolumn{1}{c|}{S}&
 \textbf{{48.29}} &
  \multicolumn{1}{c|}{\textbf{{20.14}}} &
  \multicolumn{2}{c}{\underline{{29.91}}} &
  \multicolumn{2}{c}{\textbf{{3.78}}} &
  \multicolumn{2}{c}{\underline{{4.28}}} &
  \multicolumn{2}{c}{14.34} &
  \multicolumn{2}{c}{\underline{{9.08}}} &
  \multicolumn{2}{c}{\underline{{6.65}}} &
  \multicolumn{2}{c}{\underline{{62.21}}} &
  \multicolumn{2}{c}{\textbf{{18.16}}} &
  \multicolumn{2}{c}{\underline{{40.19}}} &
  \multicolumn{2}{c}{5.16} &
  \multicolumn{2}{c}{\underline{{42.68}}} &
  \multicolumn{2}{c}{\textbf{{8.83}}} &
  \multicolumn{2}{c}{\textbf{{38.84}}} &
  \multicolumn{2}{c}{\textbf{{25.50}}} &
  \multicolumn{2}{c}{\textbf{{16.60}}} &
  \multicolumn{2}{c}{\textbf{{19.14}}} &
  \multicolumn{2}{c}{\underline{{10.30}}} &
  \multicolumn{2}{c}{\underline{{6.89}}} &\\\hline
\end{tabular}%
}
\vspace{-0.2cm}
\caption{Quantitative results on SSCBench-KITTI-360 test set. \textbf{Bold} and \underline{underline} highlight the best and second-best results, respectively.}
\vspace{-0.2cm}
\label{tab:KITTI360_test}
    
\end{table*}

\begin{table}[hbt!]
\centering
\setlength\tabcolsep{8pt}\resizebox{\linewidth}{!}{%
\begin{tabular}{c|ccc|ccc|cc}
\hline
   & \multicolumn{3}{c|}{Scan Module}     & \multicolumn{3}{c|}{$\mathcal{L}_{scan}$}       &        \multicolumn{2}{c}{Metric}            \\
   Method & depth & width & height & $\mathcal{L}_{scan}^{dep}$ & $\mathcal{L}_{scan}^{wid}$ & $\mathcal{L}_{scan}^{hgt}$ & IoU  & mIoU \\
   \hline
Baseline&&&&&&&42.23&14.91\\
\hline
(a) &      \checkmark      &            &                &       &&& 46.21&  16.50       \\
(b)      &  &      \checkmark      &                &         &&& 46.06 & 16.48       \\
(c)      &            &  &      \checkmark          &         &&& 45.92 & 16.58          \\
(d)      & \checkmark & \checkmark & \checkmark&  &&& 46.09 & 16.63\\
\hline
(e)      &  &  & &   \checkmark   &&&43.04&14.99\\
(f)      &  &  & &     & \checkmark&&43.18&15.10\\
(g)      &  &  & &    &&\checkmark  &42.95&15.04\\
(h)      &  &  & &   \checkmark &\checkmark&\checkmark  &43.08&15.19\\
\hline
ScanSSC      & \checkmark & \checkmark &\checkmark &   \checkmark &\checkmark&\checkmark  &45.95&\textbf{17.12}\\
\hline
\end{tabular}%
}
\vspace{-0.2cm}
\caption{Ablation of architectural components of the ScanSSC.}
\vspace{-0.2cm}
\label{tab:ablation_components}
\end{table}
\begin{table}[ht]
\centering
\setlength\tabcolsep{8pt}\resizebox{\linewidth}{!}{%
\begin{tabular}{c|ccc|ccc|cc}
\hline
   & \multicolumn{3}{c|}{Scan Module}     & \multicolumn{3}{c|}{$\mathcal{L}_{scan}$}       &        \multicolumn{2}{c}{Metric}            \\
   Method & depth & width & height & $\mathcal{L}_{scan}^{dep}$ & $\mathcal{L}_{scan}^{wid}$ & $\mathcal{L}_{scan}^{hgt}$ & IoU  & mIoU \\
\hline
(a) &      $\leftrightarrows$      &            &                &       &&&45.73&16.73        \\
(b)      &  &      $\leftrightarrows$      &                &         &&&45.36&16.99       \\
(c)      &            &  &      $\leftrightarrows$          &         &&& 45.94&16.96        \\
\hline
(e)      &  &  & &   $\leftrightarrows$   &&&46.18&16.51\\
(f)      &  &  & &     & $\leftrightarrows$&&46.28&16.09\\
(g)      &  &  & &    &&$\leftrightarrows$ &45.35&16.38\\
   \hline
ScanSSC&&&&&&&45.95&\textbf{17.12}\\
\hline
\end{tabular}%
}
\vspace{-0.2cm}
\caption{Ablation of axis-wise refinement directions for the Scan Module and Scan Loss. $\leftrightarrows$ denotes a flip along the axis.}
\label{tab:ablation_direction}
\vspace{-.6cm}
\end{table}

\noindent\textbf{Ablation Study of Refinement Direction.}
In Sec.~\ref{sec:method_pre}, we define the `near-to-far' directions as front-to-back, center-to-side, and top-to-bottom for the depth, width, and height axes, respectively.
Here, we conduct an ablation study in these directions by reversing the orientation of each subcomponent in the Scan Module and Scan Loss to demonstrate the effectiveness of the near-to-far refinement strategy.
As shown in Tab.~\ref{tab:ablation_direction}, reversing the directions of all subcomponents results in significant reductions in mIoU scores.
This result demonstrates the importance of the refinement direction in the proposed methods.
Between the Scan Module and Scan Loss, the Scan Loss has a greater impact on performance, leading to a more significant degradation in mIoU scores.
This outcome contrasts with the ablation study on architectural components, where using Scan Loss alone has a relatively smaller impact on performance.
As the impact of Scan Loss grows, we conclude that incorporating both components highlights the growing importance of propagating rich contextual signals in the appropriate direction, \ie towards the distant voxels.

\begin{figure}[ht]
    \centering
    \begin{subfigure}{0.35\linewidth}
        \subfloat[Depth-axis]{\includegraphics[width=\linewidth]{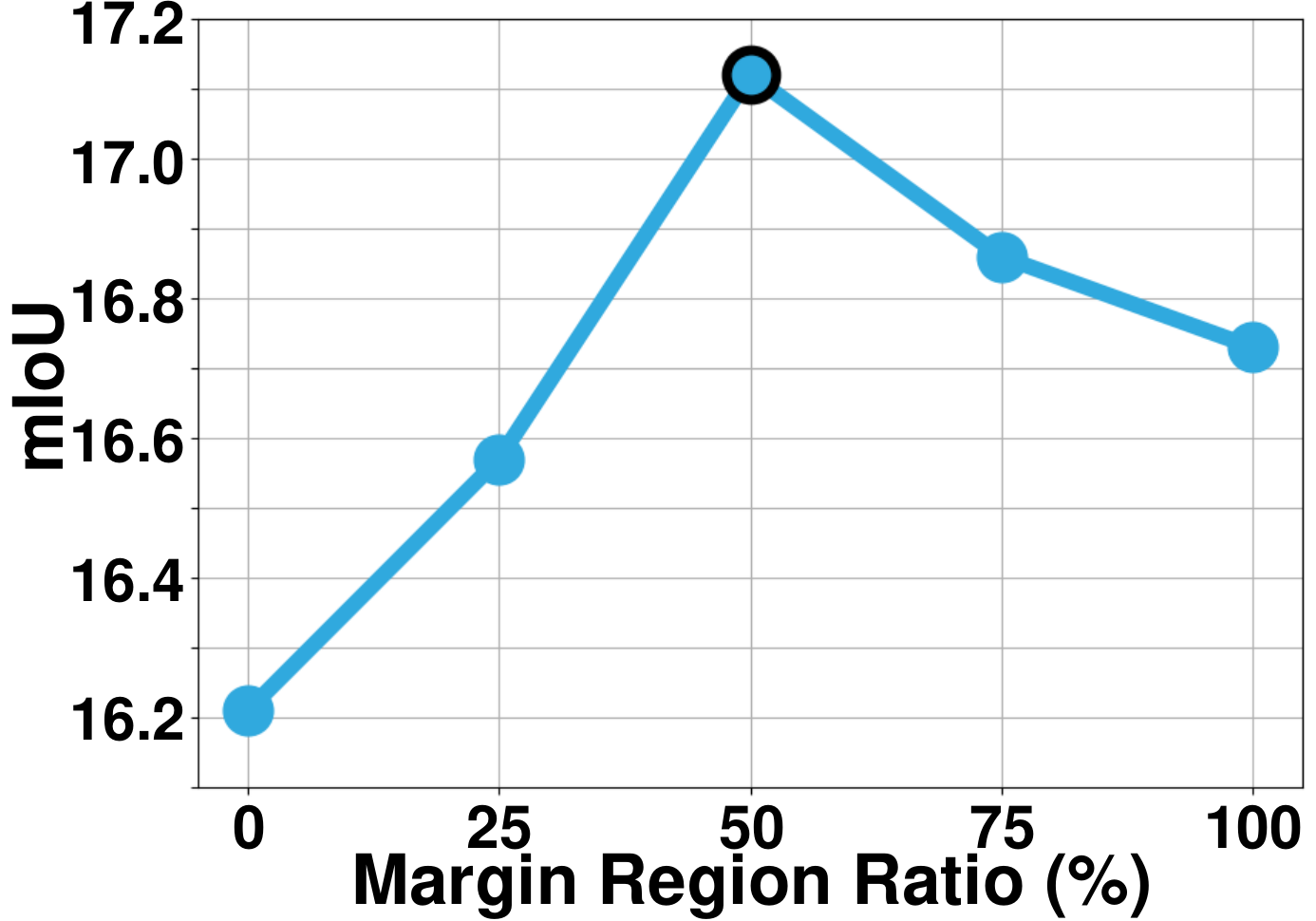}}
        \label{fig:depth_mask_ratio}
    \end{subfigure}
    \hspace{0.00\linewidth}
    \begin{subfigure}{0.305\linewidth}
        \subfloat[Width-axis]{\includegraphics[width=\linewidth, height=2.04cm]{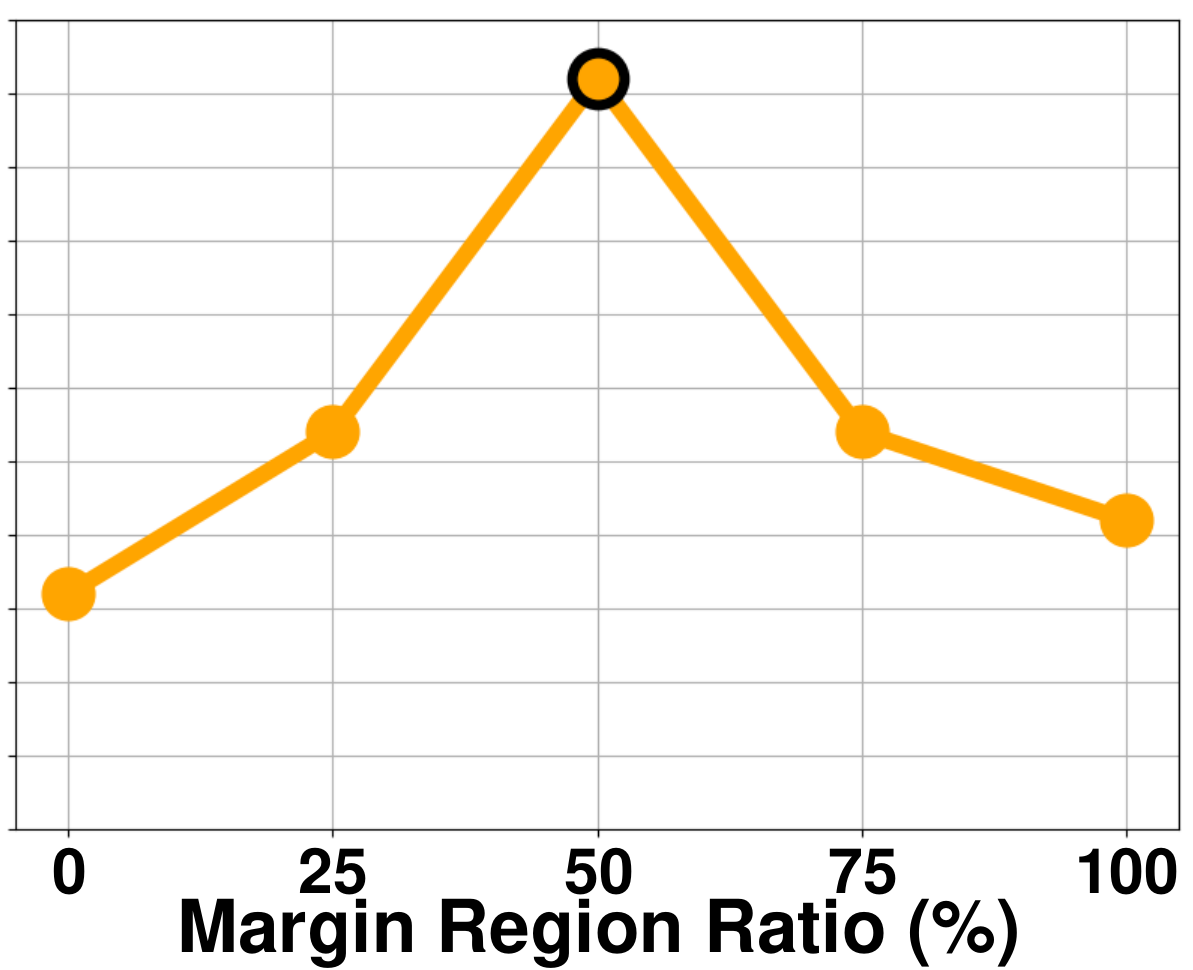}}
        \label{fig:width_mask_ratio}
    \end{subfigure}
    \hspace{0.00\linewidth}
    \begin{subfigure}{0.305\linewidth}
        \subfloat[Height-axis]{\includegraphics[width=\linewidth, height=2.04cm]{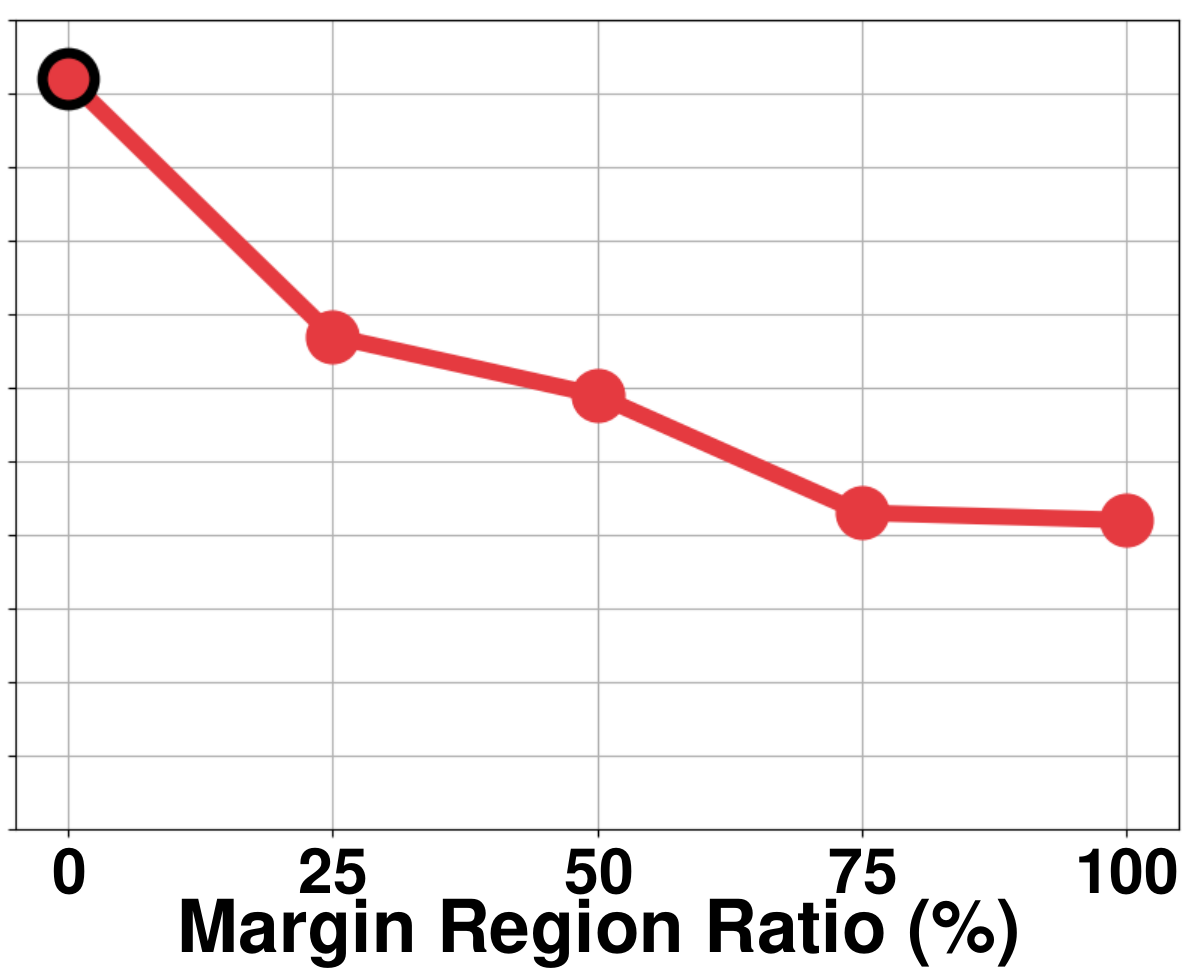}}
        \label{fig:height_mask_ratio}
    \end{subfigure}
    
\vspace{-.2cm}
    \caption{Ablation of margin region ratio in the Scan Module. }
    \label{fig:mask_ratio}
    \vspace{-.2cm}
\end{figure}

\begin{figure*}[ht]
    \centering
    \includegraphics[width=1\linewidth]{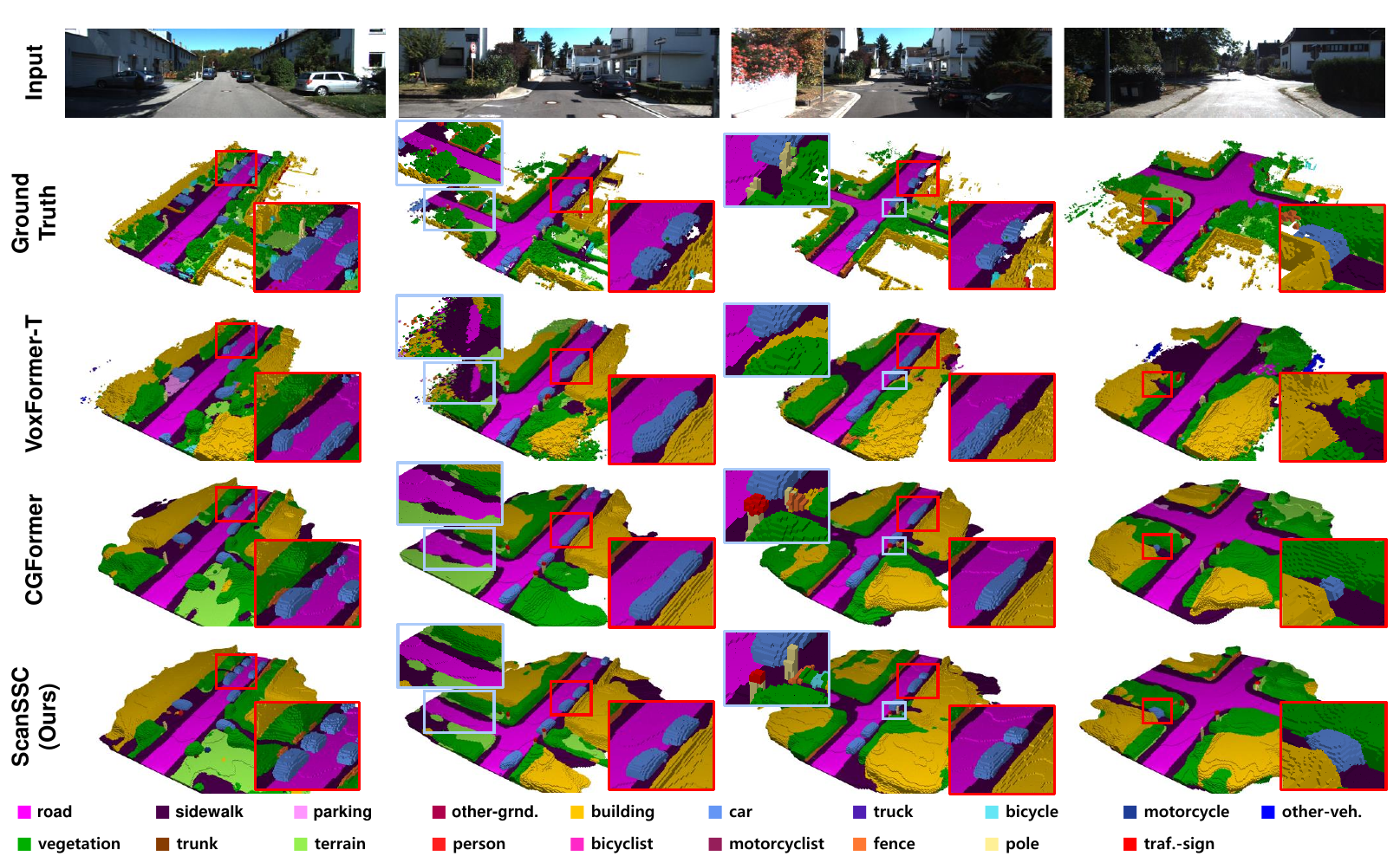}
    \vspace{-0.6cm}
    \caption{Visualization results for VoxFormer-T~\cite{li2023voxformer}, CGFormer~\cite{yu2024contextgeometryawarevoxel}, and ScanSSC on the SemanticKITTI~\cite{behley2019semantickitti} validation set.}
    \label{fig:qualitative}

\vspace{-0.6cm}
\end{figure*}

\noindent\textbf{Ablation Study of Margin Region Ratio of Masks.}
To validate the effectiveness of the margin region ratio in the Scan Module for each axis, we conduct an ablation study, with the results shown in Fig.~\ref{fig:mask_ratio}.
Each graph presents the trend of mIoU scores as the margin region ratio in the axis-specific masked self-attention varies, with the ratios for the other axes remain fixed at their default settings.
The results show that deviations from the default margin region ratios lead to a performance decrease, underscoring the importance of selecting an appropriate margin region ratio for the near-to-far strategy.
As expected, we identify that an appropriate range of interactions between near-viewpoint voxels benefits performance, while avoiding interference with inaccurate distant voxels is crucial.
In terms of height, however, it is essential to prevent the large proportion of empty voxels at higher levels from biasing the limited number of occupied voxels toward misclassification as empty.

\subsection{Qualitative Results}
\label{sec:qualitative}
We provide a visual comparison of ScanSSC with VoxFormer~\cite{li2023voxformer}, a milestone in camera-based SSC, and CGFormer~\cite{yu2024contextgeometryawarevoxel}, the current SOTA method using stereo inputs, in Fig.~\ref{fig:qualitative}.
Overall, our ScanSSC yields more plausible scene completions compared to the other methods.
The result in column 1 highlights ScanSSC's strong performance in distant scenes, as it is the only method to fully reconstruct the sequence of three cars at the end of the road.
In column 2, we observe that ScanSSC successfully reconstructs the most plausible side road areas, which are otherwise invisible due to the view frustum.
We infer that these results are primarily driven by the near-to-far refinements along the depth and width axes of ScanSSC.
Meanwhile, in columns 2 and 3, ScanSSC demonstrates its accuracy by effectively distinguishing sequentially grounded objects on the road, including cars, and smaller objects, such as poles.
Additionally, the result in column 4 shows the robustness of ScanSSC to occlusion, as it successfully reconstructs a car, which is even invisible in the input image.
At the same time, in contrast, other methods fail to capture it completely.
We believe that these results are influenced by the height-axis refinement, which provides an overview of the entire scene in BEV.
In summary, this visual analysis demonstrates the superiority of ScanSSC, highlighting its robust performance across varying distances from the viewpoint, as well as in occluded and hidden areas.

\section{Conclusion}
\label{sec:conclusion}
In this work, we first address the underestimation problem of distant scenes in existing camera-based SSC methods.
We propose ScanSSC, a novel camera-based model comprising the Scan Module and Scan Loss, based on a comprehensive analysis of prior approaches.
Both modules are designed to improve the reconstruction of distant scenes by leveraging contextual cues from near-viewpoint scenes. 
Utilizing the synergy between these components, ScanSSC achieves SOTA performance on two major SSC datasets, surpassing previous models in generating visually plausible completions for distant and occluded areas.
We believe that our study highlights key challenges in camera-based SSC that future research should address, establishing a clear path for the field.
We hope this work contributes to broader tasks in 3D computer vision and autonomous driving.

\section{Acknowledgement}
\begin{sloppypar}
This work was supported by the National Research Foundation of Korea (NRF) grant funded by the Korea government (MSIT) (No. 2023R1A2C200337911 and No. RS-2023-00220762).
\end{sloppypar}

\clearpage
\renewcommand{\thesection}{\Alph{section}}
\renewcommand{\thetable}{\Alph{section}.\arabic{table}}
\renewcommand{\thefigure}{\Alph{section}.\arabic{figure}}
\renewcommand{\thealgorithm}{\Alph{section}.\arabic{algorithm}}
\setcounter{section}{0}
\setcounter{page}{1}
\maketitlesupplementary

\section{Dataset and Metric}
\noindent\textbf{Dataset.} We evaluate ScanSSC on SemanticKITTI~\cite{behley2019semantickitti} and SSCBench-KITTI-360~\cite{li2024sscbenchlargescale3dsemantic} datasets.
SemanticKITTI is derived from the KITTI Odometry~\cite{geiger2012we} benchmark, consisting of 22 outdoor scenes
captured by LiDAR scans and stereo images.
These 22 scenes are split into 10 training scenes, 1 validation scene, and 11 test scenes.
The ground truth voxel grids have dimensions of 256$\times$256$\times$32, with each voxel measuring (0.2m, 0.2m, 0.2m), annotated with
21 semantic classes (19 semantics, 1 empty and 1 unknown). 
SSCBench-KITTI-360 is extracted from the KITTI-360~\cite{liao2022kitti}, comprising 7 training scenes, 1 validation scene, and 1 test scene.
It includes 19 semantic classes (18 semantics and 1 free).\\

\noindent\textbf{Metric.} Following standard practices in related works~\cite{li2023voxformer,jiang2024symphonize,yu2024contextgeometryawarevoxel,li2024hierarchical}, we use the mean Intersection over Union (mIoU) to assess the overall performance of semantic scene completion (SSC) and Intersection over Union (IoU) to evaluate the performance of semantic-agnostic scene completion.

\setcounter{table}{0}
\setcounter{figure}{0}
\section{More Details}
\noindent\textbf{Implementation Details.} We train ScanSSC for 25 epochs on 4 NVIDIA A6000 GPUs with a batch size of 4.
The AdamW optimizer~\cite{loshchilov2017decoupled} is used with $\beta_1=0.9$, $\beta_2=0.99$, and a maximum learning rate of $3 \times 10^{-4}$.
For the learning rate schedule, we employ a multi-step scheduler, reducing the learning rate by a factor of 0.1 at the 20\textsuperscript{th} epoch.\\

\noindent\textbf{Architectural Details.} Similar to related works~\cite{cao2022monoscene,huang2023tri,wei2023surroundocc,yu2024contextgeometryawarevoxel}, we employ a 2D UNet image encoder built upon a pretrained EfficientNetB7~\cite{tan2020efficientnetrethinkingmodelscaling}. 
Following previous stereo-based methods~\cite{jiang2024symphonize,li2023voxformer,yu2024contextgeometryawarevoxel}, we utilize the MobileStereoNet~\cite{shamsafar2021mobilestereonetlightweightdeepnetworks} as the depth estimator. 
In the viewing transformation, we adopt the depth network from CGFormer~\cite{yu2024contextgeometryawarevoxel}, which modifies the BEVDepth~\cite{li2022bevdepthacquisitionreliabledepth}.
There are 3 deformable attention layers for cross-attention and 2 for self-attention, with 8 sampling points per reference point in both heads.
The spatial mixing network consists of 3 stages, each with 2 residual blocks~\cite{he2015deepresiduallearningimage}.

\setcounter{table}{0}
\setcounter{figure}{0}
\section{Computational Cost}
We report the computational cost of ScanSSC compared to CGFormer~\cite{yu2024contextgeometryawarevoxel} in Tab.~\ref{tab:computational_cost}.
ScanSSC shows competitive efficiency, with only a slight increase in parameters and inference time.
However, it achieves notable performance improvements, with a 0.13 increase in IoU and a 0.77 increase in mIoU on the SemanticKITTI test set, highlighting the effectiveness of our method.
\begin{table}[hbt!]
\centering
\setlength\tabcolsep{5pt}\resizebox{\linewidth}{!}{%
\begin{tabular}{c|cc|cc}
\hline
   \multicolumn{1}{c|}{Method}  &     Params (M)  & Inference Time (ms) & IoU & mIoU  \\
    \hline
    CGFormer & 122 &566&44.41&16.63\\
    ScanSSC & 145 &674&\textbf{44.54}&\textbf{17.40}\\
    \hline
\end{tabular}%
}
\caption{Comparison of computational costs with CGFormer~\cite{yu2024contextgeometryawarevoxel}. The inference time for a single sample of SemanticKITTI~\cite{behley2019semantickitti} validation set is measured on 1 NVIDIA A6000 GPU.}
\label{tab:computational_cost}
\end{table}

\setcounter{table}{0}
\setcounter{figure}{0}
\section{Additional Ablation Studies}
We provide additional ablation studies to evaluate the effectiveness of the subcomponents of ScanSSC. 
Consistent with the manuscript, all experiments are conducted on the SemanticKITTI~\cite{behley2019semantickitti} validation set.
\\

\noindent\textbf{Ablation Study of Tri-Feature Fusion Methods.}
\label{sec:sup_tri}
We perform an ablation study to demonstrate the validity of ScanSSC's tri-feature fusion method by replacing it with three alternative methods, one at a time (Tab.~\ref{tab:ablation_fusion}). 
`Concat$\rightarrow$Linear' denotes the concatenation of the three axis-specific features along the channel dimension, followed by a linear layer to directly compute the output feature.
`Average' refers to an element-wise average of the three features, while `Weighted Sum' denotes a weighted summation of the three features using voxel-wise learnable parameters $L\in\mathbb{R}^{\hat{X}\times\hat{Y}\times\hat{Z}\times 3}$.
We observe that the proposed tri-feature fusion method results in a significantly higher mIoU value than the three alternative methods, demonstrating the effectiveness of voxel-wise adaptive fusion of axis-specific features.
\begin{table}[hbt!]
\centering
\setlength\tabcolsep{15pt}\resizebox{.8\linewidth}{!}{%
\begin{tabular}{
>{\columncolor[HTML]{FFFFFF}}c |
>{\columncolor[HTML]{FFFFFF}}c 
>{\columncolor[HTML]{FFFFFF}}c }
\hline
                          Method  & IoU   & mIoU  \\ \hline
Concat $\rightarrow$ Linear & 45.90 & 16.32 \\
Average                     & 46.17 & 16.51 \\
Weighted Sum                   & 45.90 & 16.28 \\ \hline
Tri-Feature Fusion                        & 45.95 & \textbf{17.12} \\ \hline
\end{tabular}}
\caption{Ablation study on the tri-feature fusion method of ScanSSC.}
\label{tab:ablation_fusion}
\end{table}

\begin{figure}[hbt!]
    \centering
    \includegraphics[width=.6\linewidth]{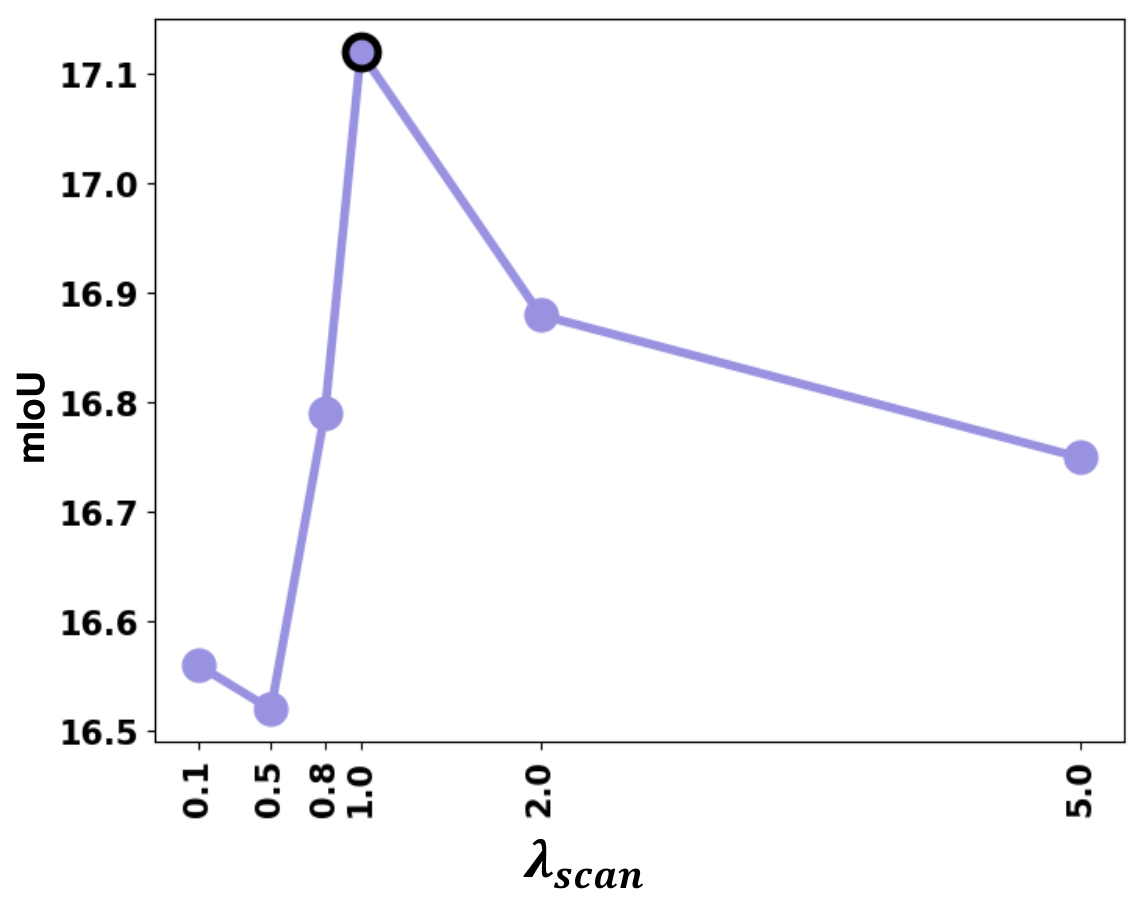}
    \caption{Performance comparison by loss scaling coefficient for Scan Loss.}
    \label{fig:lambda}

\end{figure}

\noindent\textbf{Loss Scaling Coefficient for Scan Loss.}
We conduct an ablation study on the loss scaling coefficient of the Scan Loss, $\mathcal{L}_{scan}$, as shown in Fig.~\ref{fig:lambda}.
When the coefficient is set to 1, the highest mIoU score of 17.12 is observed, while it decreases as the coefficient moves further from 1 overall.
We find that the training mIoU consistently increases proportionally with $\lambda_{scan}$ throughout the entire training.
From this result, we infer that an excessively high value of $\lambda_{scan}$ can lead to overfitting of the model, highlighting the importance of selecting an appropriate scaling coefficient.


\noindent\textbf{Scan Loss \vs General Cross-Entropy.}
Since the proposed Scan Loss is equivalent to the cross-entropy~\cite{zhang2018generalized} of the cumulatively averaged logit, incorporating it can be seen as analogous to either modifying the distribution of the loss coefficient for the existing voxel-wise cross-entropy or simply increasing its scale.
Hence, we compare the performance of ScanSSC when it is substituted with a simple coefficient adjustment for the voxel-wise cross-entropy loss, as shown in Tab.~\ref{tab:ablation_crossentropy}.
For (a), to assign higher weights to distant voxels, we first generate a bilinearly interpolated weight along each axis, ranging from 0 to 1 in the corresponding near-to-far direction, then use the average of these weights as the coefficient for the existing cross-entropy loss ($\lambda_{tri}$).
For (b), we simply apply a higher scalar weight ($\lambda_{ce}$) to the voxel-wise cross-entropy loss.
Here, we set $\lambda_{ce}$ to 10, as the converged loss scales of both methods are similar.

\begin{table}[h]
\centering
\setlength\tabcolsep{10pt}\resizebox{.7\linewidth}{!}{%
\begin{tabular}{l|cc}
\hline
   \multicolumn{1}{c|}{Method}   &   IoU  & mIoU      \\
   \hline 
\rowcolor[HTML]{FFFFFF} 
(a) $\lambda_{tri}$$\mathcal{L}_{ce}$     & \multicolumn{1}{c}{\cellcolor[HTML]{FFFFFF}46.15} & 16.50 \\
(b) $\lambda_{ce}$$\mathcal{L}_{ce}$& 45.04 & 16.74  \\
\hline
   $\mathcal{L}_{ce}$ + $\mathcal{L}_{scan}$ (Ours)& 45.95 & \textbf{17.12} \\
\hline
\end{tabular}%
}
\caption{Performance comparison between incorporating Scan Loss and adjusting the coefficient for voxel-wise cross-entropy loss~\cite{zhang2018generalized}.
The other training losses remain unchanged during training. }
\label{tab:ablation_crossentropy}
\end{table}
Comparing with (a), we observe that Scan Loss significantly enhances mIoU by leveraging the semantic distribution of previous voxels to transmit signals to the target voxel. 
This demonstrates that instead of merely assigning higher weights to distant voxels, utilizing the semantic distribution of previous voxels to propagate signals more effectively is a superior approach.
In addition, when compared to (b), the result demonstrates that the simple increase in the weight of the existing cross-entropy does not lead to performance improvement, as it results in significantly lower IoU and mIoU scores.
\\

\noindent\textbf{Ablation Study of Subsidiary Components.}
We conduct additional experiments to validate the importance of the subsidiary components, the spatial mixing network, and tri-feature fusion.
Since tri-feature fusion can not be removed entirely, we replace it with "Concat$\rightarrow$Linear" which is represented in Tab.~\ref{tab:ablation_fusion}. 
As shown in Tab.~\ref{tab:ablation_subsidiary}, the spatial mixing network and tri-feature fusion contribute to performance improvement, demonstrating that enhancing regional spatial patterns and adaptively fusing features are effective.
\begin{table}[h]
\centering
\resizebox{\columnwidth}{!}{%
\begin{tabular}{
>{\columncolor[HTML]{FFFFFF}}c |
>{\columncolor[HTML]{FFFFFF}}c 
>{\columncolor[HTML]{FFFFFF}}l |
>{\columncolor[HTML]{FFFFFF}}l 
>{\columncolor[HTML]{FFFFFF}}l }
\hline
Method  & \multicolumn{1}{l}{\cellcolor[HTML]{FFFFFF}Spatial-mixing net.} & Tri-feature fusion                                      & IoU   & mIoU  \\ \hline
(a) & \multicolumn{1}{l}{\cellcolor[HTML]{FFFFFF}} &                                                         & 44.91 & 15.90 \\
(b) & \checkmark                                   &                                                         & 45.90 & 16.32 \\
(c) &                                              & \multicolumn{1}{c|}{\cellcolor[HTML]{FFFFFF}\checkmark} & 45.08 & 16.11 \\ \hline
ScanSSC & \checkmark                                                         & \multicolumn{1}{c|}{\cellcolor[HTML]{FFFFFF}\checkmark} & \textbf{45.95} & \textbf{17.12} \\ \hline
\end{tabular}%
}
\caption{Ablation study on the subsidiary components of the ScanSSC.}
\label{tab:ablation_subsidiary}
\end{table}

\noindent\textbf{Ablation Study of Using Tri-Axes Features.}
\begin{figure*}[ht]
    \centering
    \includegraphics[width=1\linewidth]{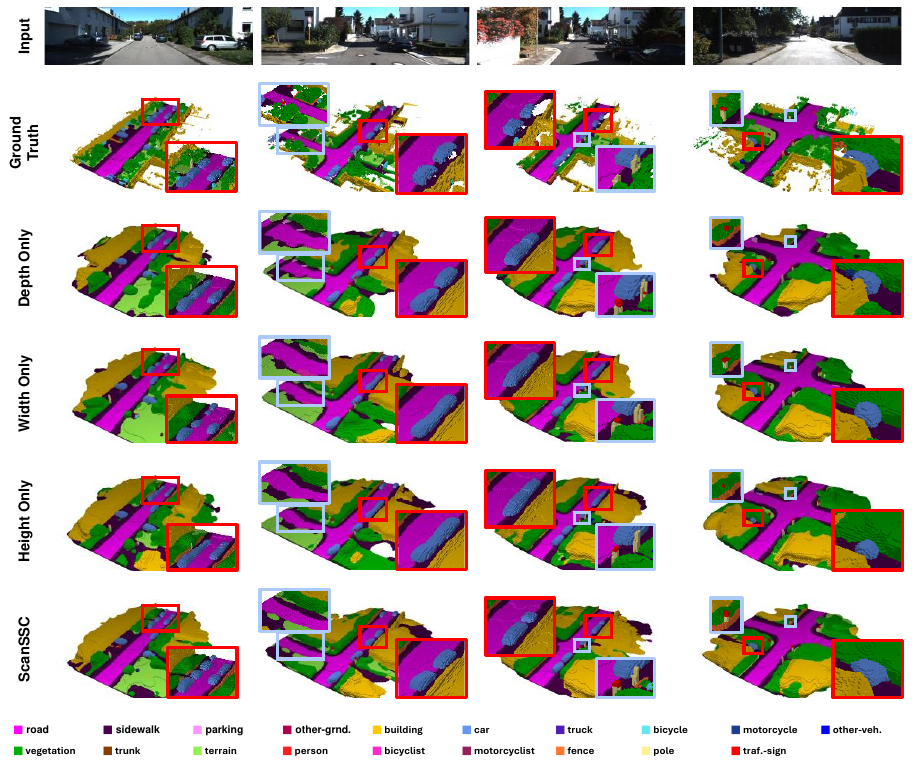}
    \caption{Visualization and comparison of the results of applying the Scan Module and Scan Loss to each individual axis and ScanSSC on the SemanticKITTI~\cite{behley2019semantickitti} validation set.}
    \label{fig:comparison_block}

\vspace{-0.4cm}
\end{figure*}

To demonstrate the validity of using features from all three axes, we visualize and compare the results obtained using features from each axis with those obtained through ScanSSC.
The results obtained using each axis individually are represented as `Depth Only,' `Width Only,' and `Height Only,' and are shown in Fig.~\ref{fig:comparison_block}.
Overall, ScanSSC, which utilizes features from all three axes, demonstrates significantly more plausible results. 
Using features from only a single axis tends to result in inaccurate predictions for distant vehicles, side road areas, and occluded objects. 
In contrast, ScanSSC, which combines features from all three axes, achieves significantly more reliable and accurate predictions. We hypothesize that this improvement arises from the complementary nature of features from each axis, which together enable a more comprehensive understanding of the entire scene.
\\



\setcounter{table}{0}
\setcounter{figure}{0}
\section{Analysis}

\noindent\textbf{Quantitative Result for Distant Geometry.}
This study aims to improve the overall reconstruction in distant regions.
To support this, quantitative analysis results are presented in Tab.2 of the manuscript.
Additionally, for a more detailed analysis, Tab.~\ref{tab:rebuttal_subcategory} provides group-wise mIoU for distant $1/2$ regions along each axis ($1/4$ on both sides for the width axis), following the categorization from the official SemanticKITTI~\cite{behley2019semantickitti} website.

\begin{table}[hbt!]
\centering
\setlength\tabcolsep{5pt}\resizebox{\linewidth}{!}{%
\begin{tabular}{
>{\columncolor[HTML]{FFFFFF}}l |
>{\columncolor[HTML]{FFFFFF}}c |
>{\columncolor[HTML]{FFFFFF}}c 
>{\columncolor[HTML]{FFFFFF}}c 
>{\columncolor[HTML]{FFFFFF}}c 
>{\columncolor[HTML]{FFFFFF}}c |
>{\columncolor[HTML]{FFFFFF}}c 
>{\columncolor[HTML]{FFFFFF}}c 
>{\columncolor[HTML]{FFFFFF}}c 
>{\columncolor[HTML]{FFFFFF}}c }
\hline
&& \multicolumn{4}{c|}{Large class group} & \multicolumn{4}{c}{Small class group} \\
                          Method & Axis & Ground   & Structure & Nature & Total & Vehicle & Human & Object & Total \\ \hline
CGFormer&Dep.&24.87&17.66&\textbf{19.57}&21.98&6.92&0.07&2.86&3.95\\
ScanSSC&Dep.&\textbf{26.29}&\textbf{17.89}&19.23&\textbf{22.60}&\textbf{7.03}&\textbf{0.19}&\textbf{3.19}&\textbf{4.11}\\ \hline
CGFormer&Wid.&14.23&\textbf{15.81}&\textbf{16.51}&15.28&1.24&0.15&1.35&0.97\\
ScanSSC&Wid.&\textbf{16.26}&14.65&15.96&\textbf{15.95}&\textbf{1.39}&\textbf{0.57}&\textbf{1.49}&\textbf{1.19}\\ \hline
CGFormer&Hgt.&29.90&\textbf{23.15}&25.73&27.49&13.23&2.79&6.73&8.61\\
ScanSSC&Hgt.&\textbf{31.69}&21.50&\textbf{25.91}&\textbf{28.25}&\textbf{13.78}&\textbf{3.44}&\textbf{7.11}&\textbf{9.14}\\ \hline

\end{tabular}}
\caption{Per-group mIoUs on the SemanticKITTI~\cite{behley2019semantickitti} validation set.}
\vspace{-.6cm}
\label{tab:rebuttal_subcategory}
\end{table}
This result highlights ScanSSC's effectiveness in distant regions, demonstrating its superior performance on both large and small geometries, particularly outperforming CGFormer~\cite{yu2024contextgeometryawarevoxel} in all small class groups.
\\

\begin{figure*}[t]
    \centering
    \includegraphics[width=1\linewidth]{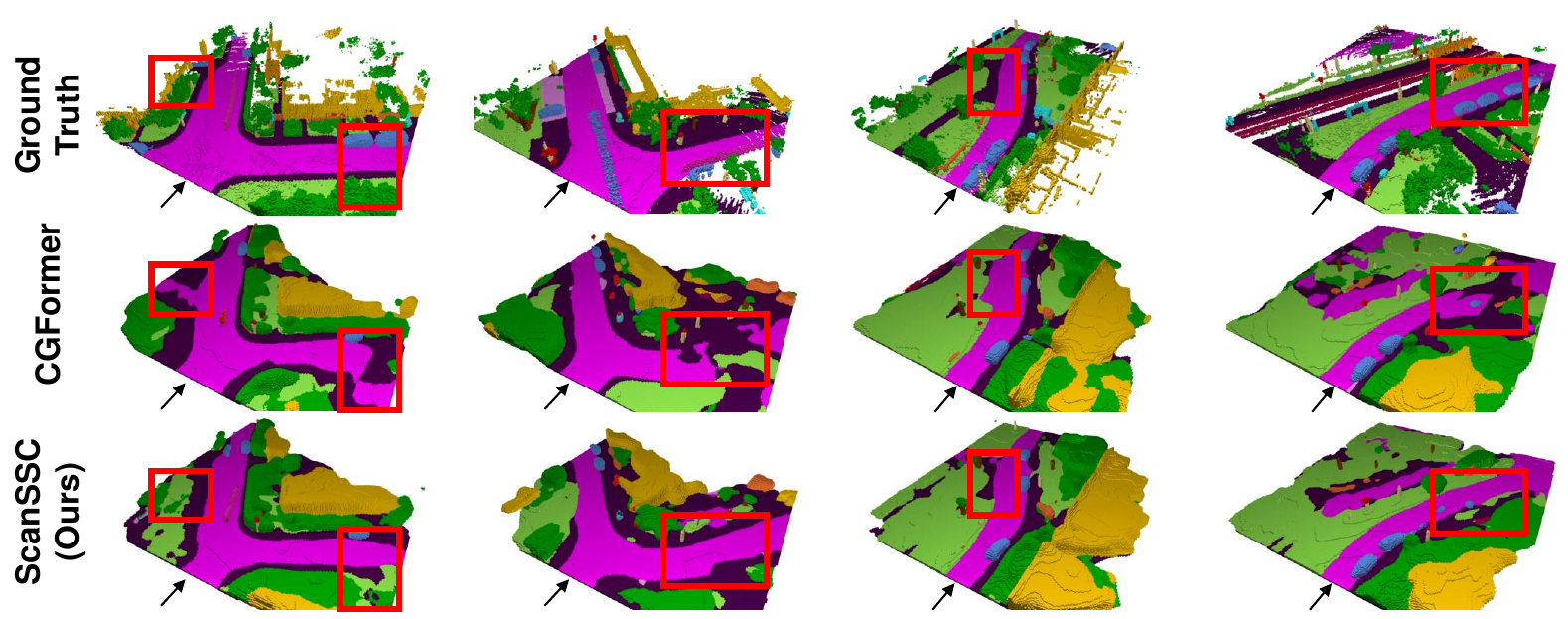}
    \caption{Visualization results of the non-axis-aligned cases of CGFormer~\cite{yu2024contextgeometryawarevoxel} and ScanSSC on the SemanticKITTI~\cite{behley2019semantickitti} validation set.}
    \label{fig:analysis_curved}

\vspace{-0.3cm}
\end{figure*}

\begin{figure*}[t]
    \centering
    \includegraphics[width=1\linewidth]{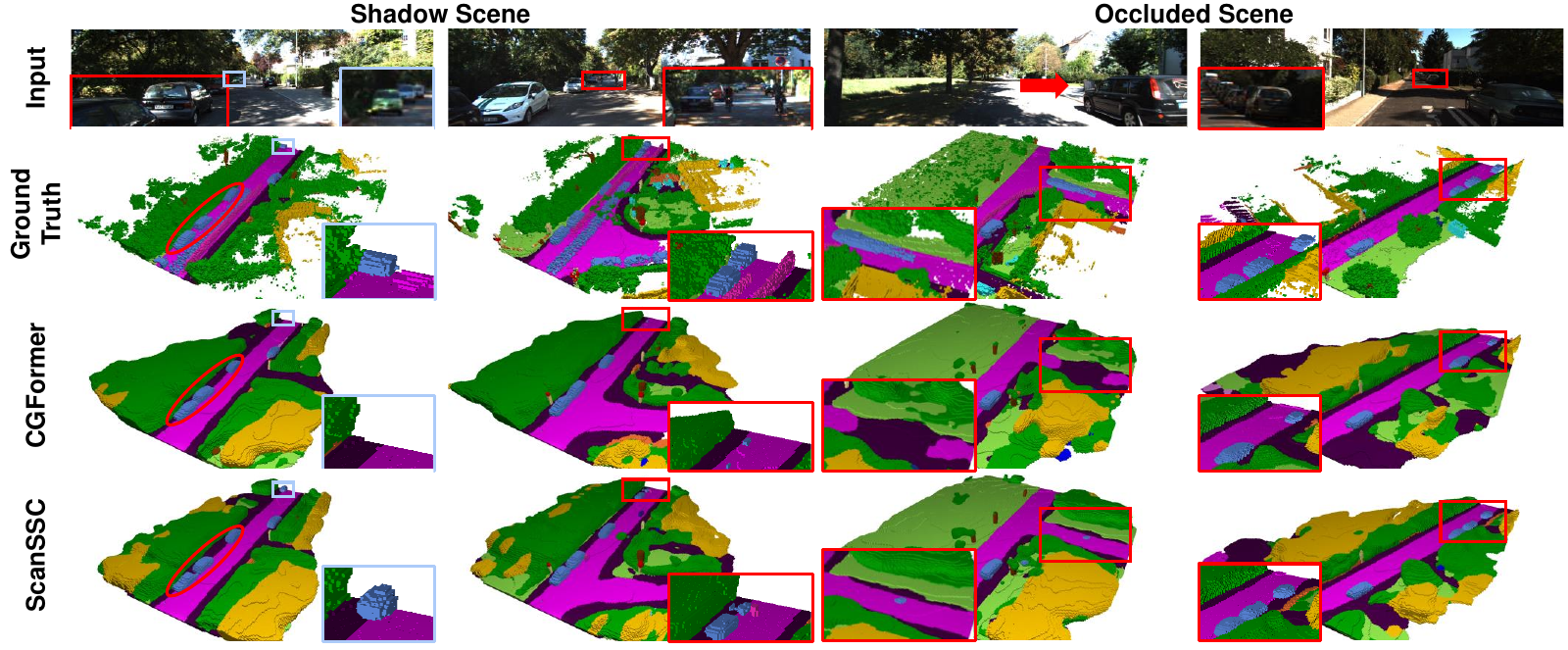}
    \caption{Visualization results of CGFormer~\cite{yu2024contextgeometryawarevoxel} and ScanSSC under various conditions (e.g., shadow, occlusion) on the SemanticKITTI~\cite{behley2019semantickitti} validation set.}
    \label{fig:analysis_conditions}

\vspace{-0.5cm}
\end{figure*}

\vspace{-0.4cm}
\noindent\textbf{Analysis of the Non-Axis-Aligned Cases.}
Since the proposed Scan Module and Scan Loss operate axis-wise, ScanSSC is effective in most axis-aligned driving scenarios, as demonstrated by the qualitative results in the manuscript.
However, this raises the question of whether ScanSSC's operation might be less effective in non-axis-aligned scenes.
To investigate this, we conduct additional evaluations of ScanSSC in non-axis-aligned scenarios.
Since the SemanticKITTI benchmark dataset does not explicitly categorize curve road scenes, we manually classify these cases.
Numerically, ScanSSC significantly outperforms CGFormer, achieving a mIoU of 14.56 and an IoU of 42.38, compared to CGFormer's 13.77 and 41.96.
As shown in Fig.~\ref{fig:analysis_curved}, ScanSSC performs comparably overall without side effects. Specifically, its performance is on par for small objects; however, it reconstructs roads significantly better in distant regions.

\noindent\textbf{Analysis of the Various Scene Conditions.}
We conduct additional analyses to demonstrate the superiority of ScanSSC under various conditions (e.g., shadow, occlusion).
Since existing benchmark datasets for SSC do not categorize various environments, we manually filter shady and highly occluded scenarios from the RGB images of the SemanticKITTI dataset.

As shown in Fig.~\ref{fig:analysis_conditions}, in the shady scenario, ScanSSC clearly distinguishes both nearby and distant vehicles covered by shadows, whereas CGFormer fails to do so. 
In addition, in the occluded scenario, ScanSSC effectively reconstructs the right-side road obscured by nearby vehicles and accurately identifies distant cars that are partially occluded. 
These results demonstrate ScanSSC's robustness across diverse and challenging scenes.

\vspace{-.2cm}
\setcounter{table}{0}
\setcounter{figure}{0}
\section{Additional Qualitative Results}
\begin{figure*}[ht]
    \centering
    \begin{subfigure}{0.24\linewidth}
        \subfloat[VoxFormer~\cite{li2023voxformer}]{\includegraphics[width=\linewidth]{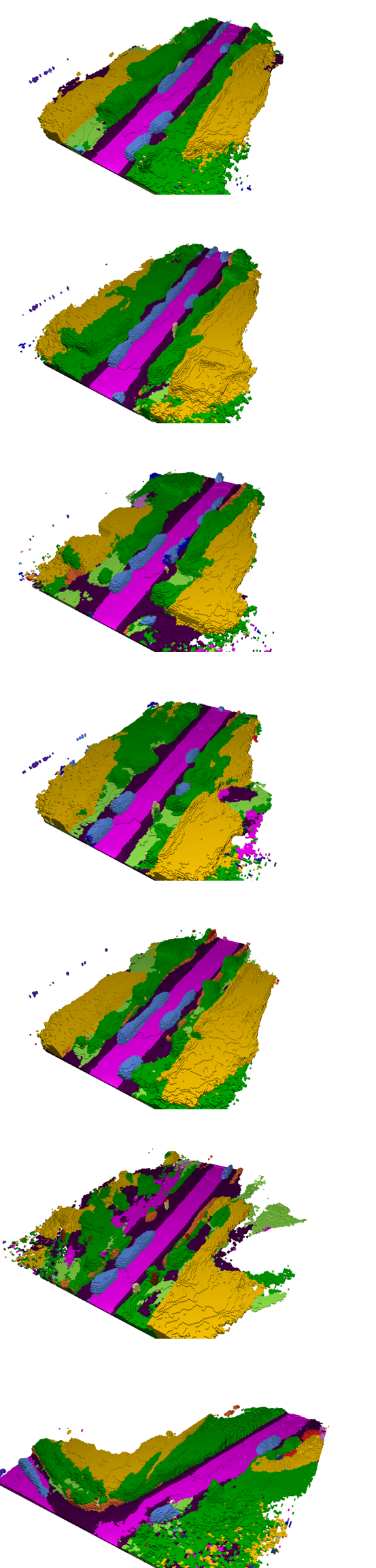}}
        \label{fig:voxformer_appen}
    \end{subfigure}
    \hspace{0.00\linewidth}
    \begin{subfigure}{0.24\linewidth}
        \subfloat[CGFormer~\cite{yu2024contextgeometryawarevoxel}]{\includegraphics[width=\linewidth]{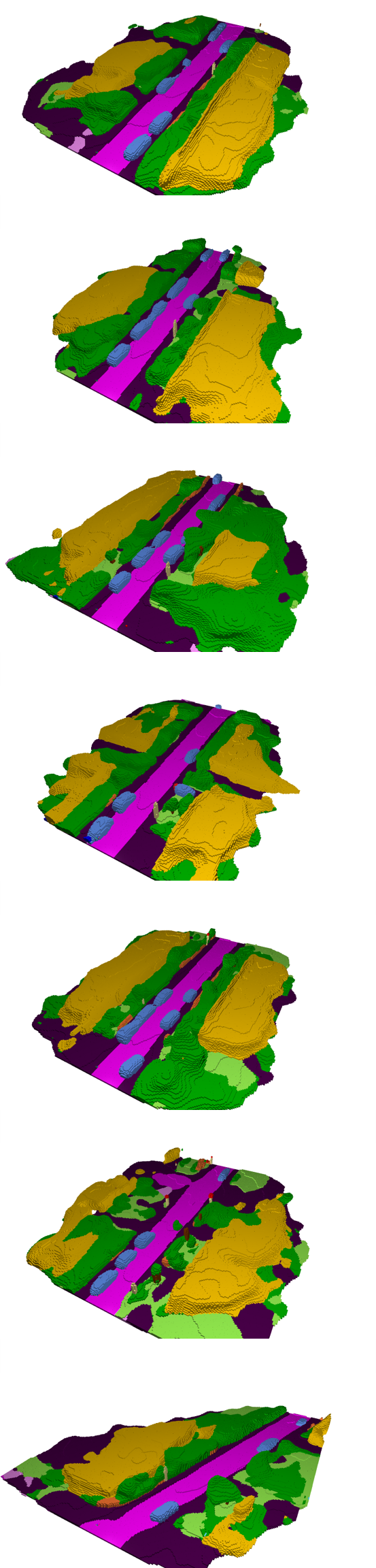}}
        \label{fig:cgformer_appen}
    \end{subfigure}
    \hspace{0.00\linewidth}
    \begin{subfigure}{0.24\linewidth}
        \subfloat[ScanSSC (Ours)]{\includegraphics[width=\linewidth]{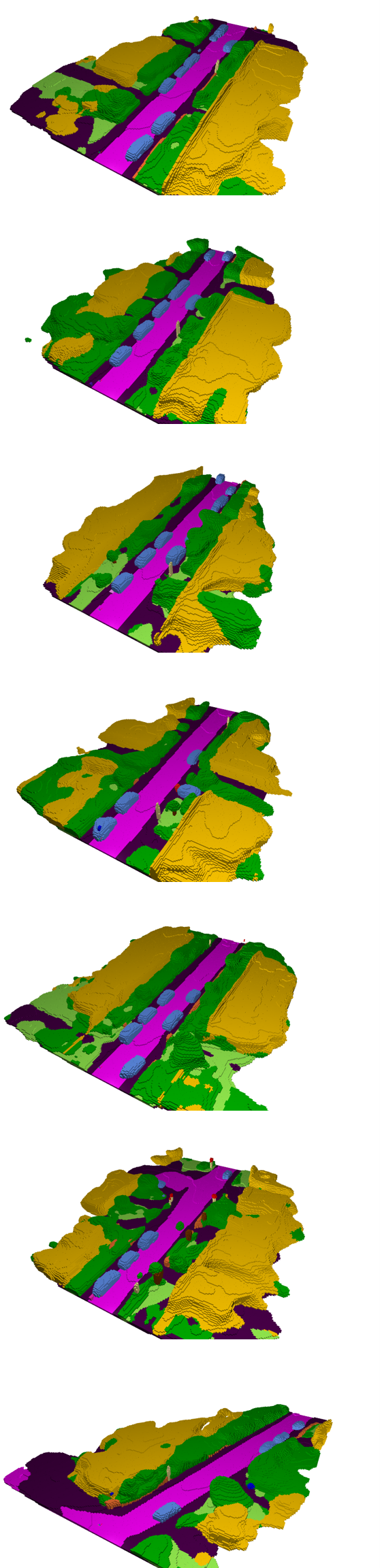}}
        \label{fig:scanssc_appen}
    \end{subfigure}
    \begin{subfigure}{0.24\linewidth}
        \subfloat[Ground Truth]{\includegraphics[width=\linewidth]{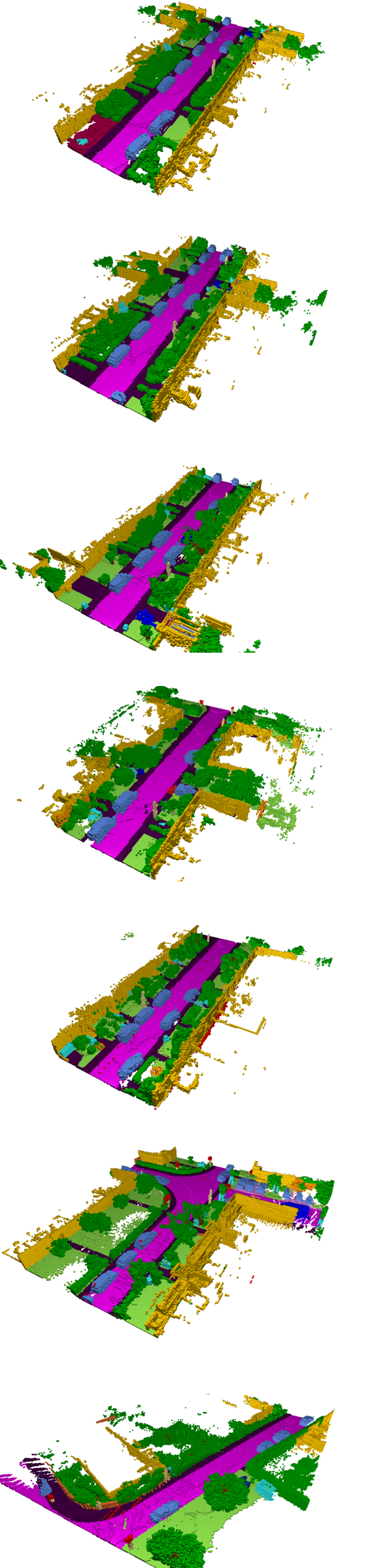}}
        \label{fig:gt_appen}
    \end{subfigure}
    
    \caption{More qualitative comparison results on the SemanticKITTI~\cite{behley2019semantickitti} validation set.}
    \label{fig:appen_comparison}
\end{figure*}
We present additional qualitative comparisons with VoxFormer~\cite{li2023voxformer} and CGFormer~\cite{yu2024contextgeometryawarevoxel}, as visualized in Fig.~\ref{fig:appen_comparison}. These results are randomly selected from the SemanticKITTI~\cite{behley2019semantickitti} validation set.



\setcounter{table}{0}
\setcounter{figure}{0}
\section{Pytorch-like Pseudocode of Scan Module and Scan Loss}
To facilitate a comprehensive understanding of the proposed Scan Module and Scan Loss, we present the PyTorch-like~\cite{paszke2019pytorch} pseudocode for each in Algorithm \ref{alg:module} and Algorithm \ref{alg:loss}, respectively.

\lstset{
  basicstyle=\ttfamily\scriptsize,
  keywordstyle=\color{blue},
  commentstyle=\color{gray},
  stringstyle=\color{orange},
  breaklines=true,
  breakindent=0pt,
  frame=single,
  language=Python,
  tabsize=2,
  showstringspaces=false
}

\begin{algorithm}[hbt!]
\scriptsize
\caption{PyTorch Style Pseudocode of Scan Module.}
\begin{lstlisting}
import torch
import torch.nn as nn

class ScanModule(nn.Module):
  def __init__(self, dim):
    # declare axis-specific Scan Blocks
    self.dep_block = ScanBlock(dim)
    self.wid_block = ScanBlock(dim)
    self.hgt_block = ScanBlock(dim)

  def forward(self, x):
    X_, Y_, Z_, C = x.size()
    x_dep = x.permute(1, 2, 0, 3).flatten(0,1)
    x_wid = x.permute(0, 2, 1, 3).flatten(0,1)
    x_hgt = x.flatten(0,1)

    # axis-specific masks
    dep_attn_mask = torch.triu(torch.ones(X_, X_), diagonal=1)==1
    dep_attn_mask[:, :X_//2] = False  # depth-axis margin region

    wid_attn_mask = torch.tril(torch.ones(Y_//2, Y_//2), diagonal=-1)==1
    wid_attn_mask = torch.cat((wid_attn_mask, wid_attn_mask.flip(dim=[-1])), dim=-1)
    wid_attn_mask = torch.cat((wid_attn_mask, wid_attn_mask.flip(dims=[-2])), dim=0)
    wid_attn_mask[:, Y_//4:-(Y_//4)] = False  # width-axis margin region

    hgt_attn_mask = torch.tril(torch.ones(Z_), diagonal=-1)==1

    # axis-wise voxel scanning
    x_dep = self.dep_block(x_dep, dep_attn_mask)
    x_wid = self.wid_block(x_wid, wid_attn_mask)
    x_hgt = self.hgt_block(x_hgt, hgt_attn_mask)

    x_dep = x_dep.reshape(Y_, Z_, X_, C).permute(2, 0, 1, 3)
    x_wid = x_wid.reshape(X_, Z_, Y_, C).permute(0, 2, 1, 3)
    x_hgt = x_hgt.reshape(X_, Y_, Z_, C)

    return x_dep, x_wid, x_hgt

class ScanBlock(nn.Module):
  def __init__(self, dim):
    self.norm1 = nn.LayerNorm(dim)
    self.masked_sa = nn.MultiheadAttention(dim)
    self.norm2 = nn.LayerNorm(dim)
    self.ff1 = nn.Linear(dim, dim*2)
    self.activation = nn.ReLU()
    self.ff2 = nn.Linear(dim*2, dim)

  def forward(self, x, attn_mask):
    B_, L_, C = x.size()
    
    # Masked Self-Attention
    x_norm1 = self.norm1(x)
    x = x + self.masked_sa(x_norm1, x_norm1, x_norm1, 
                            attn_mask = attn_mask)
    
    # Feed Forward Network                            
    x_norm2 = self.norm2(x)
    x = x + self.ff2(self.activation(self.ff1(x_norm2)))

    return x
\end{lstlisting}
\label{alg:module}
\end{algorithm}






  
\begin{algorithm}[hbt!]
\scriptsize
\caption{PyTorch Style Pseudocode of $\mathcal{L}_{\text{scan}}$.}
\begin{lstlisting}
import torch
import torch.nn.functional as F

def ScanLoss(logit, target):
  P, X_, Y_, Z_ = logit.size()
  # back to front
  cum_x = torch.cumsum(logit.flip((1)), axis=1)
  # sides to center
  cum_y_l = torch.cumsum(logit[:Y_//2], axis=2)
  cum_y_r = torch.cumsum(logit[Y_//2:].flip((2)), axis=2)
  cum_y = torch.cat([cum_y_l, cum_y_r], dim=2)
  # bottom to top
  cum_z = torch.cumsum(logit, axis=3)

  # to logit value scaling
  cum_x /= torch.arange(1, X_+1)
  cum_y /= torch.arange(1, Y_+1)
  cum_z /= torch.arange(1, Z_+1)

  # same with logits
  X_, Y_, Z_ = target.size()
  target = F.one_hot(target).permute(3, 0, 1, 2)
  cum_x_t = torch.cumsum(target.flip((1)), dim=1)
  cum_y_l_t = torch.cumsum(target[:Y_//2], dim=2)
  cum_y_r_t = torch.cumsum(target[Y_//2:].flip((2)), dim=2)
  cum_y_t = torch.cat([cum_y_l_t, cum_y_r_t], axis=2)
  cum_z_t = torch.cumsum(target, dim=3)

  cum_x_t /= torch.arange(1, X_+1)
  cum_y_t /= torch.arange(1, Y_+1)
  cum_z_t /= torch.arange(1, Z_+1)

  L_scan_x = F.cross_entropy(cum_x, cum_x_t, reduction='mean')
  L_scan_y = F.cross_entropy(cum_y, cum_y_t, reduction='mean')
  L_scan_z = F.cross_entropy(cum_z, cum_z_t, reduction='mean')
  L_scan = L_scan_x + L_scan_y + L_scan_z

  return L_scan
\end{lstlisting}
\label{alg:loss}
\end{algorithm}
\clearpage
{
    \small
    \bibliographystyle{ieeenat_fullname}
    \bibliography{supp}
}
\end{document}